\pdfoutput=1
\documentclass[10pt,twocolumn,letterpaper]{article}

\usepackage[pagenumbers]{paper}

\definecolor{linkblue}{rgb}{0.21,0.49,0.74}
\usepackage[pagebackref,breaklinks,colorlinks,allcolors=linkblue]{hyperref}

\usepackage{pifont}
\usepackage{placeins}
\usepackage{tabularx}
\usepackage{makecell}
\usepackage{array}
\usepackage{bm}
\refstepcounter{figure}
\usepackage{graphicx}
\usepackage{booktabs}
\usepackage{multirow}
\usepackage{cuted}
\usepackage{stfloats}
\usepackage{float}
\usepackage[table]{xcolor}
\definecolor{bestblue}{HTML}{D6EAFF}
\definecolor{secondgreen}{HTML}{DDF2DD}
\usepackage{capt-of}
\usepackage{pgfplots}
\pgfplotsset{compat=1.18}

\newcommand{\bestbf}[1]{\cellcolor{bestblue}\textbf{#1}}

\title{Edit4D: Editable 4D Scene Generation with Textual Control}

\author{
Chanwoo Kim$^{1,2}$ \qquad Tackgeun You$^{1}$ \qquad Hwasup Lim$^{1,2,\dagger}$\\[4pt]
{\normalsize $^{1}$Korea Institute of Science and Technology (KIST), Republic of Korea}\\
{\normalsize $^{2}$University of Science and Technology (UST), Republic of Korea}\\[4pt]
{\tt\small \{sky9739, tackgeun.you, hslim\}@kist.re.kr}\\[4pt]
{\small $^{\dagger}$Corresponding author}\\[2pt]
{\small Project page: \url{https://chanwo0kim.github.io/Edit4D/}}
}

\begin{document}
\twocolumn[{
\maketitle
\begin{center}
\includegraphics[width=0.99\linewidth]{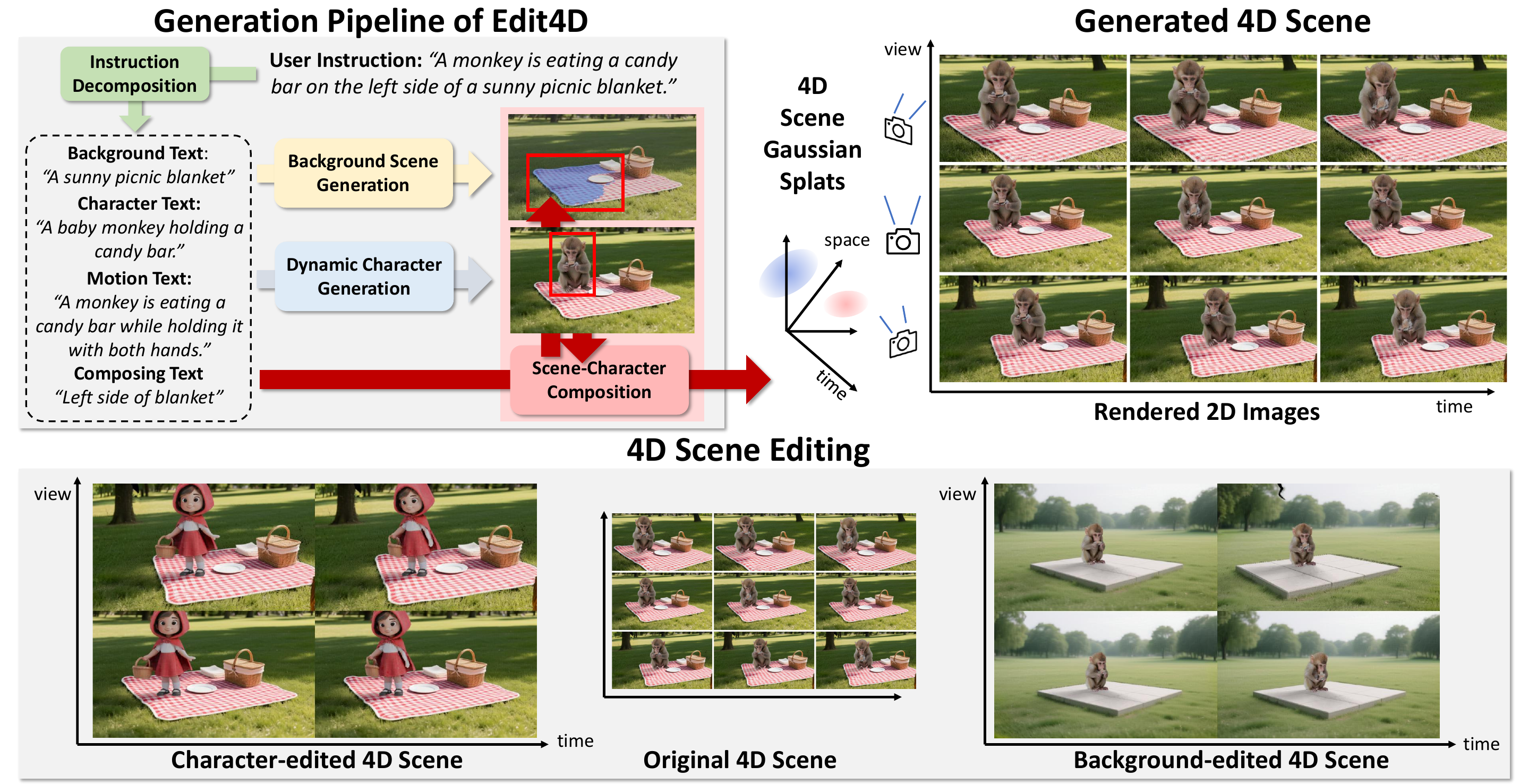}
\label{fig:teaser}
\parbox{\linewidth}{
\small
\textbf{Figure 1.} Overview of Edit4D.
Given a user instruction, Edit4D first decomposes it into stage-specific prompts.
Each stage of the pipeline then takes the corresponding prompts for the scene, character appearance, and motion, and generates the corresponding Gaussian splatting for each stage.
We further propose a composition stage that naturally places the character into the background and renders them jointly.
By constructing the scene from separated background and dynamic components, Edit4D enables flexible and independent editing.
}
\end{center}
}]
\begin{abstract}

We introduce Edit4D, an editable text-to-4D scene generation framework that constructs 4D scenes from separate background and dynamic character representations.
Recent advances in generative models have enabled text-driven dynamic scene synthesis, but editable 4D scene generation remains challenging due to the need to preserve spatial geometry and temporal dynamics.
Existing approaches typically generate holistic scene videos, where the background and dynamic characters are coupled, making object-level editing and flexible camera control difficult.
Edit4D addresses these limitations through a stage-specific generation pipeline.
Given a user instruction, Edit4D first decomposes it into stage-specific prompts for background, dynamic character, and scene--character composition.
Using these prompts, we synthesize multi-view scene images with text-to-image and camera-conditioned generative priors and reconstruct a background 3D Gaussian splatting representation.
We then generate a character image, animate it with an image-conditioned video prior, reconstruct a 4D mesh, and bind Gaussian primitives to obtain a character representation.
Finally, both components are composed through text-guided segmentation and contact-aware optimization. 
A global scene refinement is then performed to improve harmonization and rendering consistency.
By integrating stage-specific components, Edit4D enables controllable camera motion, flexible object-level editing, and the practical construction of editable 4D scenes.
Project page: \url{https://chanwo0kim.github.io/Edit4D/}
\end{abstract}
\section{Introduction}
\label{sec:intro}
Advances in large-scale text-to-image models such as DALL-E~\cite{DALL-E} and Stable Diffusion~\cite{LDM} have rapidly accelerated visual content generation.
This progress is especially evident in video generation, where models such as Sora~\cite{liu2024sora} and Veo~\cite{GoogleDeepMind2025Veo3} demonstrate strong photorealism, temporal coherence, and scene complexity.
Building on these advances, recent research has shifted toward 3D and 4D content creation.
In particular, 4D generation aims to capture both spatial structure and temporal evolution, requiring multi-view consistency, temporal coherence, and stable 3D geometry.
Despite progress in diffusion models, general 4D scene generation remains challenging and has yet to reach the level required for widespread commercial use.

Existing text-to-4D methods can be broadly divided into object-level and scene-level approaches.
Object-level methods~\cite{bahmani20244dfy, zheng2024unified} often produce multi-view consistent dynamic subjects by focusing on a single foreground object, but they do not generate complete scenes with backgrounds.
Scene-level methods~\cite{yu20244real,liu2025free4d} synthesize complete dynamic scenes as a single holistic representation, making object-level editing difficult. 
Moreover, because geometry is reconstructed from a limited set of generated viewpoints, rendering quality often degrades under novel-view changes.
These challenges are further amplified by the scarcity of large-scale dynamic-scene datasets, which forces many methods to rely on distillation from image or video generative priors. 
Although such priors provide useful supervision, the resulting 4D content remains constrained by their quality and failure modes. 
Furthermore, when artifacts appear in the generated content, even localized errors require regenerating the entire sequence, making correction costly and prone to inconsistencies with the original output.
In real production workflows, where expert-driven refinement is often necessary, this lack of editability becomes a significant limitation.
These observations suggest that constructing 4D scenes as composable assets, rather than as a single monolithic video, offers a more flexible alternative.
Asset-based generation naturally aligns with graphics workflows, allowing creators to manipulate and replace scene components with greater control.
However, editable 4D scene construction remains underexplored, as existing pipelines often couple backgrounds with dynamic elements and make independent editing difficult.

To address these limitations, we introduce Edit4D, an editable text-to-4D framework that constructs 4D scenes from separate background scene and dynamic character representations.
Given a user instruction, Edit4D first decomposes it into stage-specific prompts for background, dynamic character, and composition.
We then generate the 4D scene through parallel background scene and dynamic character generation, followed by scene--character composition.
The background scene is reconstructed as a 3D Gaussian representation from multi-view images synthesized using text-to-image and camera-conditioned generative priors.
For the dynamic character, we generate a reference video by animating a character image, reconstructing a temporally consistent 4D mesh, and binding Gaussian primitives to obtain a dynamic character representation.
The composition stage localizes the target placement region with text-guided masks and optimizes the character pose using contact-aware constraints. 
We further perform a global scene refinement to improve appearance consistency and rendering details.
Extensive experiments show that Edit4D achieves state-of-the-art performance against recent 4D generation methods, while additionally supporting controllable camera motion.
We illustrate the overview of our method in Fig.~\ref{fig:teaser}.

\begin{figure*}[!ht]
    \centering
    \includegraphics[width=\linewidth]{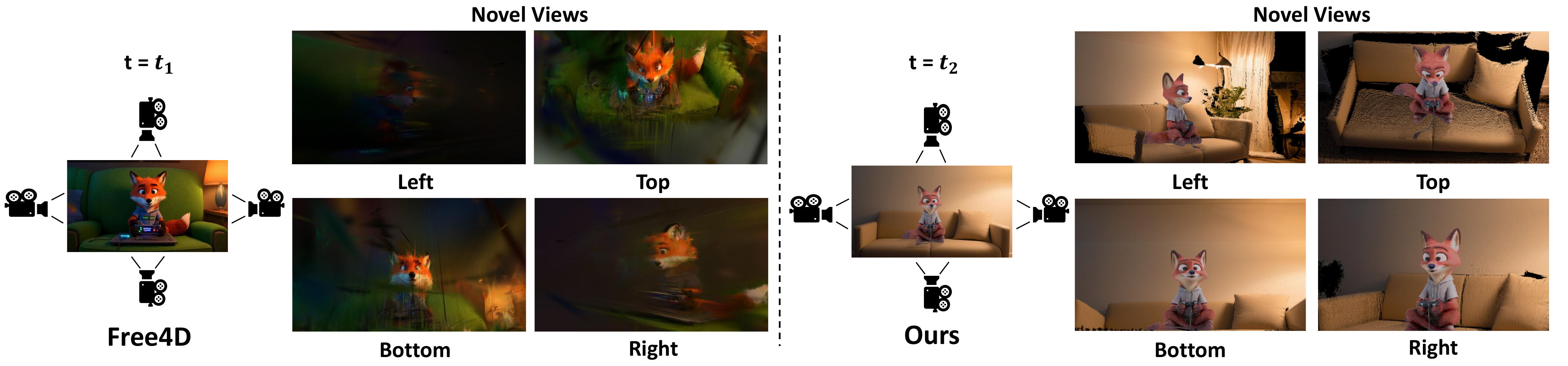}
    \vspace{-1em}
    \caption{Comparison of novel-view renderings from 4D scenes generated by Free4D~\cite{liu2025free4d} and our method.
    Compared to methods that directly distill video diffusion priors, our composition-based method constructs the 4D scene from explicit background and dynamic 3D components, maintaining better visual quality under novel-view camera changes.}
    \label{fig:compare_camera}
\end{figure*}

We summarize our contributions as follows:
\begin{itemize}
\item We propose Edit4D, an editable text-to-4D framework that constructs 4D scenes by composing separate background and dynamic character representations through a stage-specific generation pipeline. This component-wise design enables independent editing, flexible control, and scene recomposition within a unified 4D space.
\item We introduce a text-guided composition method that automatically places a dynamic character into a background scene. Our method localizes target region from text and optimizes character position with contact and penetration constraints, reducing the need for manual placement.
\item Edit4D achieves state-of-the-art performance compared with recent 4D generation methods while preserving high perceptual quality. Beyond generation quality, it supports camera control and fine-grained object-level editing.
\end{itemize}
\section{Related Works}
\label{sec:related}

\subsection{3D and 4D Scene Generation}
Text-to-3D generation has advanced rapidly with the development of large-scale generative models.
Score distillation sampling (SDS)-based methods~\cite{poole2022dreamfusion, wang2023prolificdreamer, chen2023fantasia3d, liang2024luciddreamer, zhuo2024vividdreamer, lin2023magic3d} synthesize 3D content by aligning rendered views with text-conditioned diffusion priors, but they are computationally expensive and often suffer from view-dependent artifacts and the Janus problem.
Recent multi-view guided methods address these limitations by generating geometrically consistent views and then lifting them into 3D representations.
Text-conditioned~\cite{li2024instant3d, shi2023mvdream, li2024dual3d, huang2025mv} and image-conditioned~\cite{voleti2024sv3d, liu2023one, liu2024one, liu2024syncdreamer, yang2024hunyuan3d} approaches follow this paradigm to improve visual fidelity and efficiency.
More recently, data-driven models~\cite{zhao2025hunyuan3d, wu2024direct3d, tang2024lgm} trained on large-scale 3D datasets generate 3D from text or image inputs.

Building on these advances, recent text- or video-to-4D methods~\cite{pan2024efficient4d, ren2023dreamgaussian4d, ren2024l4gm, singer2023text, sun2024eg4d, yin20234dgen, zheng2024unified, bahmani20244dfy, yu20244real, liu2025free4d} synthesize dynamic 3D content using diffusion guidance, multi-view generation, or SDS-based optimization.
Object-level methods such as 4Dfy~\cite{bahmani20244dfy} and Dream-in-4D~\cite{zheng2024unified} can produce temporally coherent and multi-view consistent dynamic objects.
Scene-level methods such as 4Real~\cite{yu20244real} and Free4D~\cite{liu2025free4d} instead generate complete dynamic scenes, improving photorealism or temporal consistency through video-conditioned reconstruction or point-guided denoising.
However, these methods typically perform holistic 4D synthesis, making it difficult to isolate and control individual scene components after generation.
In contrast, Edit4D constructs 4D scenes from separate representations, enabling fine-grained modification while preserving spatio-temporal coherence.

\begin{figure*}[t]
\begin{center}
\includegraphics[width=0.99\linewidth]{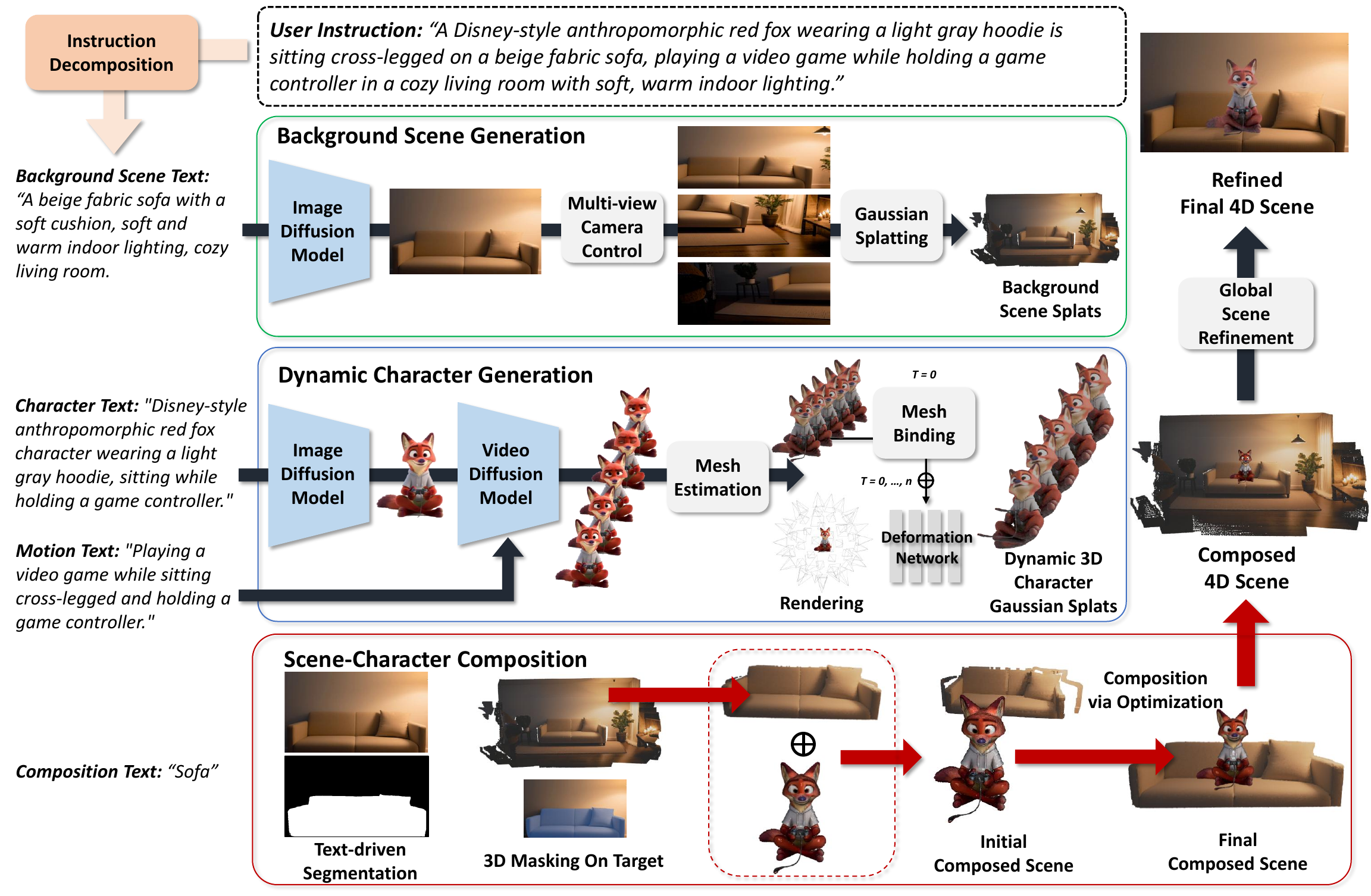}
\end{center}
\vspace{-1.5em}
\caption{\textbf{Detailed Generation Pipeline of Edit4D.}
In the first stage, Edit4D decomposes a user instruction into background scene text, character appearance and motion text, and composition text for target placement via off-the-shelf large language model.
In the second stage, our framework generates background scene and dynamic character.
A background scene is generated as a 3D Gaussian representation using text- and camera-conditioned priors.
It then constructs a dynamic 3D character via video-guided mesh reconstruction and Gaussian binding.
In the third stage, both components are composed through text-guided segmentation and contact-aware optimization to obtain an editable 4D scene.
Finally, we perform a global scene refinement to improve appearance harmonization and rendering consistency.
}
\vspace{-1em}
\label{fig:method}
\end{figure*}

\subsection{Scene Representation}
Neural Radiance Fields (NeRF)~\cite{mildenhall2021nerf} established a foundation for neural 3D scene representations through volumetric rendering.
Although NeRF achieves high visual fidelity, its slow rendering speed motivated more efficient variants such as Instant-NGP~\cite{mueller2022instant} and HexPlane~\cite{cao2023hexplane}, which leverage hash-based or tensor-decomposed encodings.
More recently, Gaussian Splatting~\cite{kerbl20233d} introduced a rasterization-based representation using Gaussian primitives, enabling real-time rendering with high visual quality.
This line of work has been extended to dynamic 4D representations, including NeRFPlayer~\cite{song2023nerfplayer}, 4D-GS~\cite{wu20244d}, and Grid4D~\cite{xu2024grid4d}, which jointly model spatial structure and temporal motion.
However, accurately recovering geometry remains challenging when optimization relies mainly on photometric supervision.
Recent methods such as SuGaR~\cite{guedon2024sugar} and 2D Gaussian~\cite{Huang2DGS2024} redesign Gaussian primitives to align with underlying surfaces, while structure-guided frameworks like GaussianAvatar~\cite{moon2024gaussianavatar, hu2024gaussianavatar} and Mani-GS~\cite{gao2025mani} attach the Gaussians to deformable meshes for stable geometry.

\subsection{Scene Composition}
With recent advances in segmentation foundation models~\cite{kirillov2023segment, zhang2024evf}, mask-based composition methods in 3D neural scenes have been actively studied. 
By leveraging 2D masks from segmentation or text-to-mask models, prior methods localize editable regions for object removal and inpainting, ensuring reliable composition~\cite{bartrum2024replaceanything3d, ye2024gaussian, shahbazi2024inserf}.
More closely related to our work, human-scene placement methods aim to place human in 3D scene with physically plausible contact while reducing interpenetration artifacts.
PROX~\cite{Hassan_2019_ICCV} fits SMPL-X~\cite{SMPL-X:2019} bodies to scenes using contact and penetration constraints, and POSA~\cite{hassan2021populating} learns a body-centric contact representation for plausible placements.
Furthermore, COINS~\cite{Zhao:ECCV:2022} and Narrator~\cite{xuan2023narrator} refine using contact optimization to encourage natural human-scene contact.
However, these methods typically assume reconstructed real scenes, supervised interaction data, or manually prepared assets.
In contrast, Edit4D performs text-driven scene--character composition, allowing dynamic characters to be automatically placed into generated background scenes through text-guided segmentation and contact-aware optimization.
\section{Method}

We propose Edit4D, a unified framework for generating editable 4D scenes from textual descriptions.
As shown in Fig.~\ref{fig:method}, our framework composes independently generated background scenes and dynamic characters into a unified 4D representation, enabling object-level editing and recomposition.
We first describe our editable 4D scene representation based on Gaussian splatting.
We then explain how Edit4D constructs a 4D scene through a stage-specific generation pipeline using off-the-shelf generative priors.

\subsection{Editable 4D Scene with Gaussian Splatting}
We represent a 4D scene using Gaussian splatting defined by the following Gaussian splatting parameters:
\begin{align}
\mathcal{G} = \left\{\left(\bm{\mu}_i, \bm{r}_i, \bm{s}_i, \alpha_i, \bm{c}_i\right)
\right\}_{i=1}^{N},
\label{eq:scene_gs_params}
\end{align}
where $\bm{\mu}$, $\bm{r}$, $\bm{s}$, $\alpha$, and $\bm{c}$ denote the 3D mean, rotation, scale, opacity, and spherical-harmonics coefficients, respectively, and $N$ is the total number of Gaussians.

Our goal is to generate a time-varying Gaussian representation conditioned on a user instruction.
Instead of modeling the entire scene as a single holistic representation, Edit4D constructs the 4D scene from separate background scene and dynamic 3D representations.
Formally, we define Gaussian splatting parameters at time $t$ as follows:
\begin{align}
\mathcal{G}(t)
=
{\mathcal{G}}^{\text{bg}}
\;\oplus\;
\left( \bigoplus_{i=1}^{N^{\text{obj}}}
\mathcal{G}^{\text{obj}}_i (t)\right),
\end{align}
where ${\mathcal{G}}^\text{bg}$ denotes the background Gaussian representation, and ${\mathcal{G}}^{\text{obj}}_i(t)$ denotes the time-dependent Gaussian representation of the $i$-th dynamic object.
The $\oplus$ operator indicates concatenation over Gaussian primitives.

This component-wise formulation naturally supports object-level editing, such as modifying the appearance, motion, or spatial placement of individual objects.
Moreover, by leveraging a standard Gaussian splatting renderer, we can freely control the camera pose and render the resulting scene from arbitrary viewpoints.

Following the Gaussian splatting formulation, we denote the differentiable renderer as $\mathcal{R}$.
Given a camera pose $\mathbf{c}_k$ at time $t$, the rendered frame is defined as:
\begin{align}
I_{\mathbf{c}_k, t}
=
\mathcal{R}\!\left(
\mathcal{G}(t)
\;,\;
\mathbf{c}_k
\right).
\end{align}
This representation supports novel-view rendering, dynamic camera control, and component-level editing.

\subsection{Stage 1: Instruction Decomposition}
Edit4D takes a single user instruction that describes the background scene, dynamic character, and scene--character composition.
Since each module requires different information, the original instruction is not directly suitable for all stages of the pipeline.
We therefore use a large language model~\cite{openai_chatgpt} to convert the user instruction into stage-specific prompts while preserving its original semantics.
As shown in Fig.~\ref{fig:method}, the decomposed prompts guide background scene generation, dynamic 3D character generation, and scene--character composition.
This instruction decomposition allows each module to operate with an appropriate prompt and enables the overall pipeline to run automatically from a single user instruction.

\subsection{Stage 2: Background Scene Generation}
\label{sec:background_scene}
To generate a background scene, we first synthesize a high-quality scene image from the background scene text. 
This image serves as a reference that defines the overall layout, appearance, and semantic structure of the scene. 
We then employ a camera-conditioned multi-view generation model to expand the image into a set of geometrically consistent views along sampled camera trajectories. 
These generated views provide additional spatial observations and parallax cues necessary for reliable 3D reconstruction.
Finally, we reconstruct a background Gaussian splatting representation from the generated multi-view observations, yielding a globally consistent background representation $\mathcal{G}^{\text{bg}}$.

\subsection{Stage 2: Dynamic 3D Character Generation}

We use a text-to-image model to generate a character image and animate it with the image-conditioned video generator.
Given the generated video, we employ a mesh estimator~\cite{chen2026motion} to reconstruct a deformable mesh sequence $\{\mathbf{M}_t\}_{t=1}^{T}$ with temporally consistent topology across time. 
To enable joint rendering with the background scene, we convert the mesh sequence into a 4D Gaussian splatting representation using the mesh-binding scheme following~\cite{gao2025mani} to exploit mesh topology.
Specifically, Gaussians are attached to mesh faces and updated according to the underlying mesh deformation, providing a topology-aware dynamic representation.

While mesh binding provides a strong motion prior, it is insufficient to model fine-grained deformations and may suffer from vertex twisting artifacts. 
We train a deformation field $f_\theta$ that predicts Gaussian offsets from the first-frame character Gaussian positions $\mathbf{x}_1$ and timestamp $t$:
\begin{align}
f_\theta(\mathbf{x}_1, t)=(\Delta \mathbf{x}, \Delta \mathbf{r}, \Delta \mathbf{s}, \Delta \mathbf{c}),
\end{align}
where $\Delta\mathbf{x}$, $\Delta\mathbf{r}$, $\Delta\mathbf{s}$, and $\Delta\mathbf{c}$ denote the offsets of position, rotation, scale, and color attributes, respectively.

For supervision, we render multi-view images $\mathbf{I}_\text{gt}^{k}\in\mathbb{R}^{H\times W\times3}$ from the mesh sequence in Blender using cameras sampled on a sphere around the subject. 
The deformation field is optimized with photometric and structural losses as follows:
\begin{align}
\mathcal{L}=|\hat{I}-I_\text{gt}|+\text{SSIM}(\hat{I},I_\text{gt}).
\end{align}

\subsection{Stage 3: Scene--Character Composition}

To reduce manual effort in composing separately generated background scenes and dynamic characters, we introduce an automated alignment procedure, as illustrated in Fig.~\ref{fig:method}.
Given a target object label (e.g., sofa or floor), we first obtain a 2D segmentation mask $M_{\text{2D}}$ using an open-vocabulary segmentation model~\cite{zhang2024evf}.
We then construct a 3D Gaussian mask by selecting Gaussians whose projections fall within $M_{\text{2D}}$.
For each Gaussian mean $\mathbf{x}$, the corresponding image coordinate $(u,v)$ is computed using the camera intrinsic matrix $\mathbf{K}$ and extrinsic parameters $(\mathbf{R},\mathbf{T})$:
\begin{align}
\mathbf{x}^{c} = \mathbf{R}\mathbf{x} + \mathbf{T},~~
\begin{bmatrix}
u \\ v \\ 1
\end{bmatrix}
=
\mathbf{K}
\begin{bmatrix}
x^{c} / z^{c} \\
y^{c} / z^{c} \\
1
\end{bmatrix},
\end{align}
where $\mathbf{x}^c = (x^c,y^c,z^c)$ denotes the point in camera coordinates.
To remove projection ambiguities, we further filter candidates using the depth map $D^{\text{bg}}$ rendered from the background scene Gaussian splatting:
\begin{align}
(u,v)\in M_{\text{2D}} ~~\text{and}~~ \left| z^c - D^{\text{bg}}(u,v) \right| < \tau,
\end{align}
where $\tau$ is a depth tolerance set to $5 \mathrm{cm}$. 
The remaining Gaussians form the masked 3D region $M_{\text{3D}}$, which serves as the target region for character composition.

We define character-object contact along the $y$-axis, consistent with the $y$-down coordinate system used by the background scene generator.
Using the masked 3D Gaussian $M_{\text{3D}}$, we compute the target object's bounding box and use it as a geometric prior for character initialization.
Specifically, the character is translated along the $-y$ direction relative to the bounding box, placing it near the target surface before optimization.
This initialization improves optimization stability and reduces failure cases caused by implausible placements, as illustrated in Fig.~\ref{fig:analysis_composition}.

To avoid implausible penetration while still allowing stable contact between instances, we introduce two complementary terms defined on the character mesh $\mathbf{M}$ and the background scene Gaussians.
Following~\cite{hasson2019learning, jiang2021hand}, the penetration loss penalizes background scene points located inside the character mesh.
By introducing the penetration loss, we prevent the contact loss from acting excessively, thereby avoiding cases where the character passes through objects and promoting plausible contact as follows:
\begin{align}
\mathcal{L}_{\text{penetration}} = \frac{1}{|\mathcal{I}|}\sum\limits_{x \in \mathcal{I}}||\mathcal{N}_{\mathbf{x}^{\mathbf{M}}}(x)-x||,
\end{align}
where $\mathcal{I}$ denotes the set of background scene points located inside the character mesh, and $\mathcal{N}_{\mathbf{x}^\textbf{M}}(x)$ denotes the nearest neighbor of $x$ among the mesh vertices $\mathbf{x}^{\mathbf{M}}$.
This term discourages interpenetration by pushing the character away from the object geometry.

Next, we introduce a nearest-neighbor contact loss with margin $\tau$ defined on the masked object region $M_{\text{3D}}$.
To avoid attracting the character toward unintended regions, we restrict the optimization to a down-sampled subset $S$ consisting of the top $\alpha\%$ mesh vertices $\mathbf{x}^{\mathbf{M}}$ that are closest to the target object along the $y$-axis.
\begin{align}
\mathcal{L}_\text{contact} = \frac{1}{|S|}\sum_{x \in S}\max(0, ||x-\mathcal{N}_{M_{\text{3D}}}(x)||-\tau).
\end{align}
Together with the penetration loss, we optimize the global translation and rotation of the character to align it with the target region in the background scene.

\subsection{Global Scene Refinement}
Although the proposed composition stage produces physically plausible scene--character composition, appearance discrepancies may still remain between the independently generated background and character.
To improve visual coherence, we perform a global scene refinement on the composed 4D scene.
Given a rendered image $I_{\mathbf{c}_k,t}$ from camera viewpoint $\mathbf{c}_k$ at time $t$, we optimize the scene using a text-conditioned diffusion prior guided by the user instruction.
Unlike the composition stage, which only updates the character pose, we jointly optimize the background scene, canonical character Gaussians, and the deformation field.
This refinement encourages appearance harmonization between independently generated components while improving fine-grained visual details.
Following the SDS-based refinement method~\cite{li2024director3d}, we optimize the rendered scene using both latent-space and image-space consistency terms:
\begin{align}
\mathcal{L}_{\text{refine}}
=
\mathbb{E}_{t,\mathbf{c}_k,\tau,\epsilon}
\Big[
\gamma(\tau)
\big(
\lambda_1 ||z_{\mathbf{c}_k,t}-\hat z_{\mathbf{c}_k,t}||_2^2
\nonumber \\+ 
\lambda_2 ||I_{\mathbf{c}_k,t}-\hat I_{\mathbf{c}_k,t}||_2^2
\big)
\Big].
\end{align}
where $z_{\mathbf{c}_k,t}=\epsilon(I_{\mathbf{c}_k,t})$ denotes the latent representation encoded by the diffusion model encoder $\epsilon$.
For the canonical time $t=0$, we sample the camera used during background scene generation to improve multi-view consistency.
For dynamics, we only use the frontal view to preserve character identity and temporal consistency during optimization.

Furthermore, we improve scene--character consistency by generating cast shadows based on the estimated scene illumination.
We estimate an HDR environment map from background multi-view renderings~\cite{liang2026luxdit} and compute character-induced light occlusion following Gaussian ray tracing~\cite{3dgrt2024,gu2024gtracer}.
The resulting attenuation is applied to the rendered background, producing temporally consistent cast shadows that follow the scene illumination.
Further implementation details are provided in Appendix~\ref{sec:supp_impl}.
\section{Experiments}

\subsection{Evaluation Metrics}

Following the evaluation protocol of Free4D~\cite{liu2025free4d}, we evaluate all methods using frontal-view videos from publicly available examples.
We use three VBench~\cite{huang2024vbench} metrics. 
First, \textbf{Text Align} measures semantic similarity between generated videos and input prompts using ViClip~\cite{wanginternvid} video-text embeddings.
\textbf{Consistency} measures temporal coherence of subjects and backgrounds using DINO~\cite{caron2021emerging} and CLIP~\cite{radford2021learning} feature similarities.
Finally, \textbf{Aesthetic} evaluates visual quality using the LAION Aesthetic Predictor~\cite{laion_aesthetic_predictor}, which considers factors such as layout, harmony, photorealism, and overall artistic impression.

\begin{table}[!t]
\small
\resizebox{0.47\textwidth}{!}{%
\begin{tabular}{llccc}
\toprule
Settings & Method & Text Align & Consistency & Aesthetic \\
\midrule
& 4Real~\cite{yu20244real} & 26.7\% & 98.1\% & 52.6\% \\
~~~(a) & Free4D~\cite{liu2025free4d} & 24.7\% & 97.6\% & 62.3\% \\
& Ours & \bestbf{27.1}\% & \bestbf{99.4}\% & \bestbf{68.3}\% \\
\midrule
& 4Dfy~\cite{bahmani20244dfy} & 27.6\% & 98.1\% & 56.2\% \\
~~~(b) & Free4D~\cite{liu2025free4d} & 25.2\% & 98.1\% & 61.2\% \\
& Ours & \bestbf{28.0}\% & \bestbf{99.2}\% & \bestbf{66.2}\% \\
\midrule
& Dream-in-4D~\cite{zheng2024unified} & 26.6\% & 97.6\% & 56.2\% \\
~~~(c) & Free4D~\cite{liu2025free4d} & 25.5\% & 97.1\% & 63.2\% \\
& Ours & \bestbf{29.2}\% & \bestbf{99.3}\% & \bestbf{74.8}\% \\
\bottomrule
\end{tabular}%
}
\vspace{-0.5em}
\caption{Quantitative comparison with previous studies. We measure text alignment, consistency, and aesthetic quality using VBench~\cite{huang2024vbench} on three experimental settings.}
\label{tab:vbench}
\vspace{-0.5em}
\end{table}

\begin{table}[t]
\centering
\tiny
\resizebox{0.38\textwidth}{!}{%
\begin{tabular}{lcc}
\toprule
Method & 3D Structure & Visual Quality \\
\midrule
Free4D~\cite{liu2025free4d} & 39.5\% & 35.0\% \\
Ours & \textbf{60.5\%} & \textbf{65.0\%} \\
\bottomrule
\end{tabular}%
}
\vspace{-0.5em}
\caption{Pairwise user study comparing Free4D and Edit4D. We report user preference for 3D structure and overall visual quality.}
\label{tab:user_study}
\vspace{-1em}
\end{table}

\subsection{Quantitative Results}
Tab.~\ref{tab:vbench} reports quantitative comparisons with prior 4D generation methods under the three evaluation settings defined by Free4D~\cite{liu2025free4d}.
Edit4D achieves state-of-the-art performance across the VBench metrics, obtaining strong text alignment, temporal consistency, and aesthetic quality.
These results demonstrate that Edit4D preserves strong generation quality while providing a more flexible and editable 4D scene representation through its stage-specific generation pipeline.

We further conduct a user study with 25 participants against Free4D, a publicly available scene level 4D generation method, using the same video generation model for a fair comparison.
Participants compare the generated results in terms of 3D Structure and Visual Quality.
As shown in Tab.~\ref{tab:user_study}, Edit4D is preferred over Free4D for both criteria, suggesting that our method provides better 3D structure and perceptual quality.
Additional details and visualizations are provided in Appendix~\ref{sec:supp_user_study}.

\begin{figure*}[!t]
\centering
\includegraphics[
    width=0.95\linewidth,
]{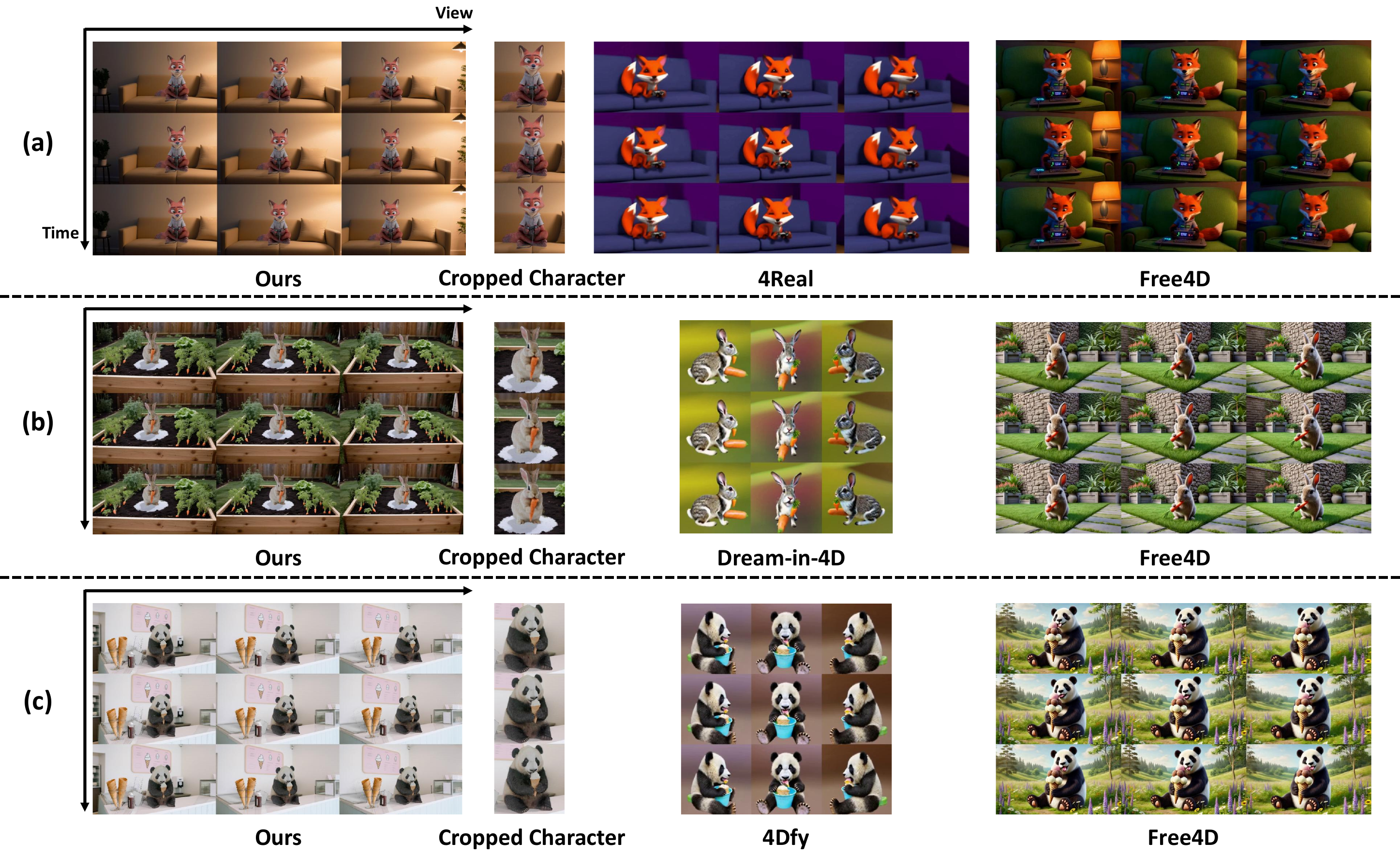}
\vspace{-0.5em}
\caption{Qualitative comparison of our method with 4Real~\cite{yu20244real}, Dream-in-4D~\cite{zheng2024unified}, 4Dfy~\cite{bahmani20244dfy}, and Free4D~\cite{liu2025free4d}. We visualize samples from experimental settings (a), (b), and (c) in Tab.~\ref{tab:vbench}. Our results show comparable visual quality, while providing editability through separated representations and the proposed composition algorithm.}
\label{fig:qualitative}
\vspace{-1em}
\end{figure*}

\begin{table*}[t]
\centering

\resizebox{0.95\textwidth}{!}{
\begin{tabular}{lccccccc|ccccccc}
\toprule
\multirow{3}{*}{Method}
& \multicolumn{7}{c|}{Scene--Character Recomposition}
& \multicolumn{7}{c}{Character Appearance Editing} \\

\cmidrule(lr){2-8}
\cmidrule(lr){9-15}

& \multicolumn{3}{c}{Success Rate $\uparrow$}
& \multicolumn{2}{c}{Preservation $\uparrow$}
& \multirow{2}{*}{Time (min) $\downarrow$}
& \multirow{2}{*}{\shortstack{User\\Study}}
& \multicolumn{3}{c}{Success Rate $\uparrow$}
& \multicolumn{2}{c}{Preservation $\uparrow$}
& \multirow{2}{*}{Time (min) $\downarrow$}
& \multirow{2}{*}{\shortstack{User\\Study}} \\

\cmidrule(lr){2-4}
\cmidrule(lr){5-6}
\cmidrule(lr){9-11}
\cmidrule(lr){12-13}

& GLM-4.6 & Qwen3-VL & Mistral-3.1 & CLIP & DINO & &
& GLM-4.6 & Qwen3-VL & Mistral-3.1 & CLIP & DINO & & \\

\midrule

Free4D
& 0\% & 40\% & 40\%
& 0.964 & 0.901
& 180
& 16.4\%
& 0\% & 40\% & 0\%
& 0.957 & 0.871
& 180
& 10.0\% \\

Ours
& \textbf{100\%} & \textbf{100\%} & \textbf{100\%}
& \textbf{0.997} & \textbf{0.997}
& \textbf{45}
& \textbf{83.6\%}
& \textbf{100\%} & \textbf{100\%} & \textbf{100\%}
& \textbf{0.991} & \textbf{0.988}
& \textbf{45}
& \textbf{90.0\%} \\

\bottomrule
\end{tabular}
}
\vspace{-0.5em}
\caption{
Quantitative evaluation of editability.
We evaluate editing success using GLM-4.6~\cite{vteam2025glm45vglm41vthinkingversatilemultimodal},
Qwen3-VL~\cite{qwen3technicalreport}, and Mistral-3.1~\cite{mistralsmall3124binstruct2503},
background preservation using CLIP~\cite{radford2021learning} and DINO~\cite{oquab2023dinov2} similarities,
as well as editing time and user preference.}
\vspace{-1.5em}
\label{tab:editability}
\end{table*}

\subsection{Qualitative Results}
\begin{figure*}[!t]
\centering
\includegraphics[
    width=0.95\linewidth
]{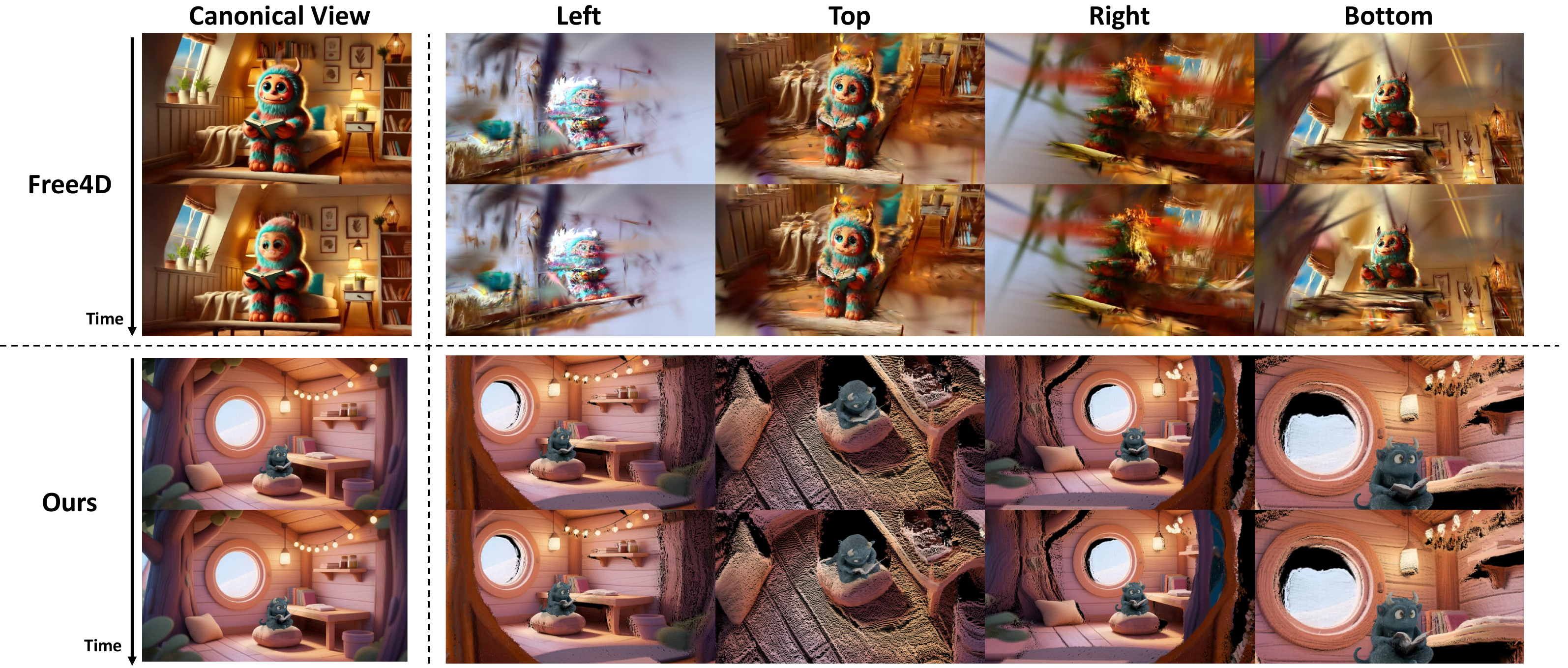}
\vspace{-1em}
\caption{Visualization of generated 4D scenes from the frontal view and four diverse novel camera views over time. Our method can preserve consistency under large camera motions, producing fewer view-dependent artifacts compared to Free4D.}
\label{fig:novel_view_rendering}
\vspace{-1em}
\end{figure*}

We compare our method with 4Real~\cite{yu20244real}, Dream-in-4D~\cite{zheng2024unified}, 4Dfy~\cite{bahmani20244dfy} and Free4D~\cite{liu2025free4d} in Fig.~\ref{fig:qualitative}.
Despite generating scene components independently, Edit4D produces coherent and natural 4D scenes with visual quality comparable to existing methods.
Moreover, as shown in Fig.~\ref{fig:novel_view_rendering}, we also render the 4D scene from four novel viewpoints over time.
Compared with the previous method~\cite{liu2025free4d}, our approach maintains more stable visual quality under large camera motions, demonstrating the effectiveness of our framework that  constructs the 4D scene from independently generated components.

\subsection{Evaluation on Editability}
\begin{figure}[!t]
    \centering
    \vspace{-0.5em}
    \includegraphics[width=0.95\linewidth]{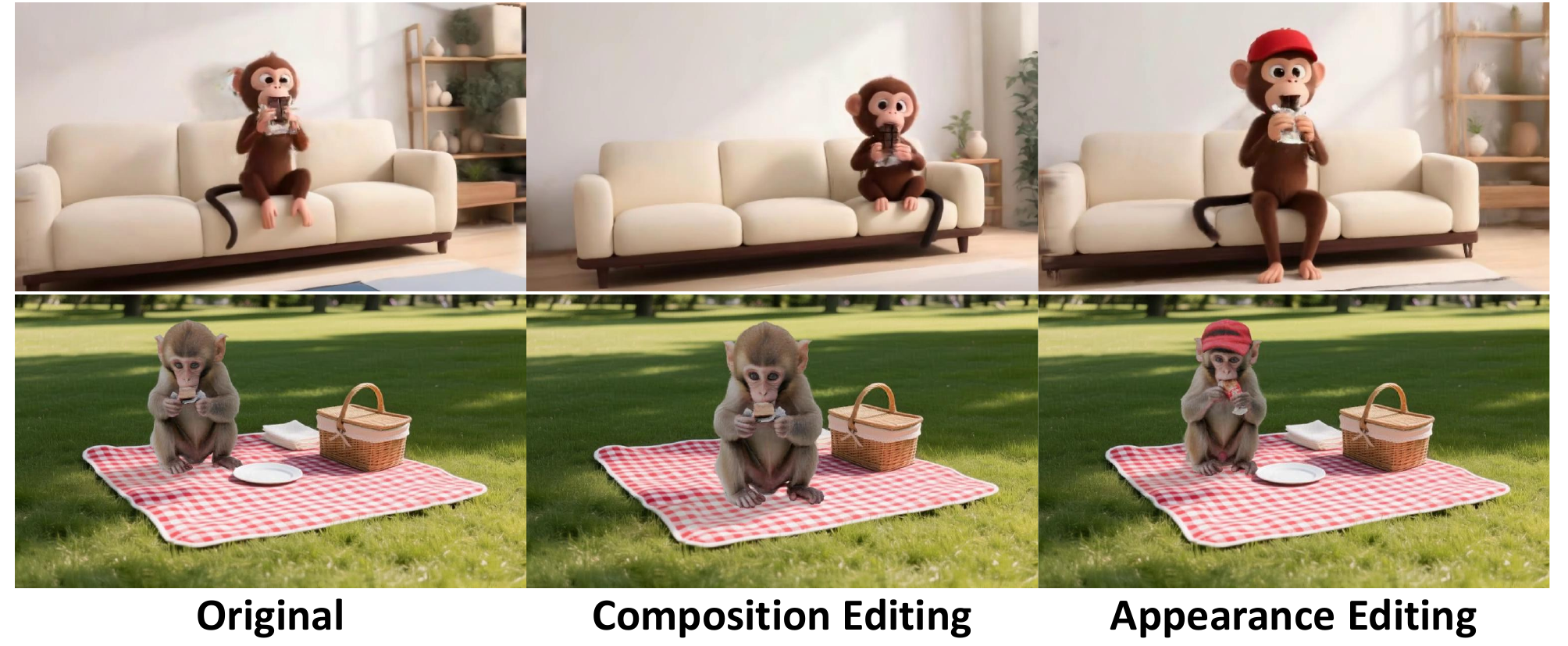}
    \vspace{-0.5em}
    \caption{Top: Free4D, bottom: Edit4D. Since Free4D requires full regeneration for editing, background content such as the bookshelf and rug changes noticeably, whereas Edit4D better preserves the original scene during composition and appearance editing.}
    \label{fig:editing_evaluation}
    \vspace{-1em}
\end{figure}

\vspace{-0.5em}
We evaluate the editability of our method through scene--character recomposition and character appearance editing.
For comparison, Free4D requires full regeneration because it represents the scene holistically. 
Therefore, we generate the edited reference videos using the same model, and execute the released Free4D code.
Since the goal of editing is to modify the target component while preserving the rest of the scene, we define editing success based on the preservation of the original background scene.
We assess this using three VLMs~\cite{vteam2025glm45vglm41vthinkingversatilemultimodal, qwen3technicalreport, mistralsmall3124binstruct2503}.
Moreover, we measure background preservation using CLIP~\cite{radford2021learning} and DINO~\cite{oquab2023dinov2} similarities after masking the character regions.
For the editing evaluation, we additionally conduct a user study to assess how well the remaining scene content is preserved after editing.
As shown in Tab.~\ref{tab:editability}, Edit4D achieves higher editing success and background preservation while reducing editing time.
In Fig.~\ref{fig:editing_evaluation}, Free4D shows noticeable changes in background content, such as the bookshelf and rug, due to full regeneration, whereas Edit4D preserves the original scene more consistently.
This efficiency comes from our asset based representation, which reuses components instead of regenerating the entire 4D scene.

\begin{table}[!t]
\resizebox{0.47\textwidth}{!}{%
\begin{tabular}{ccccccc}
\toprule
Init. & $\mathcal{L}_{\text{contact}}$ & $\mathcal{L}_{\text{penetration}}$
& \makecell{Global \\ Refinement}
& Text Align & Consistency & Aesthetic \\
\midrule
& \checkmark & \checkmark &
& 26.8\% & 98.7\% & 68.0\% \\

\checkmark & & \checkmark &
& 25.0\% & 98.9\% & 67.2\% \\

\checkmark & \checkmark &
& & 26.6\% & 98.9\% & 67.7\% \\

\checkmark & \checkmark & \checkmark &
& 26.5\% & 98.9\% & 68.2\% \\

\checkmark & \checkmark & \checkmark & \checkmark
& \bestbf{28.1}\% & \bestbf{99.4}\% & \bestbf{69.0}\% \\
\bottomrule
\end{tabular}%
}
\vspace{-0.5em}
\caption{Ablation study on the effect of each component in the scene--character composition and global scene refinement.}
\label{tab:ablation}
\vspace{-1.5em}
\end{table}

\begin{figure}[!t]
    \centering
    \vspace{-0.5em}
    \includegraphics[width=0.8\linewidth]{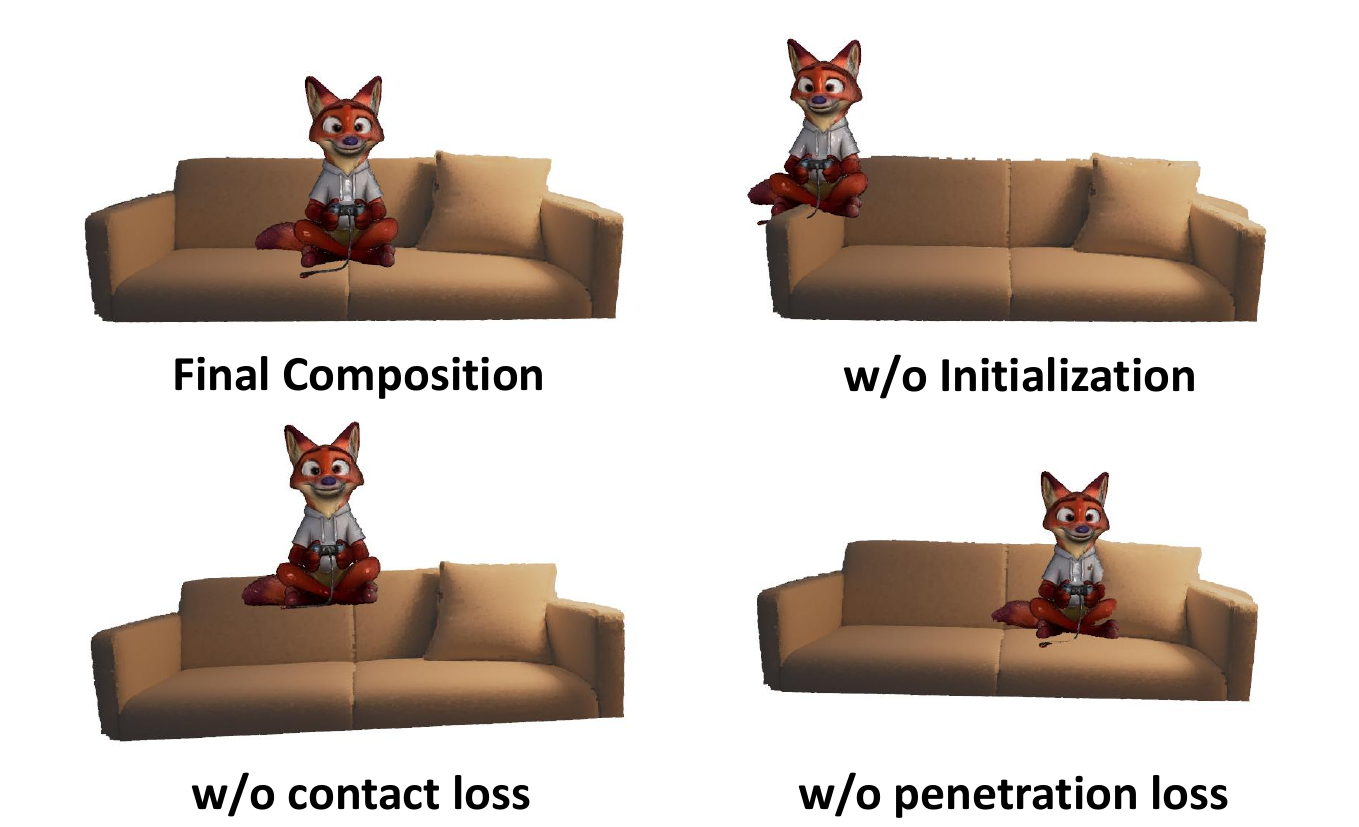}
    \vspace{-1em}
    \caption{Ablation study of our composition method. We demonstrate that our approach can reliably find the optimal placement for inserting the character based on object-specific text.}
    \label{fig:analysis_composition}
    \vspace{-2em}
\end{figure}

\begin{figure}[!t]
    \vspace{-0.5em}
    \centering
    \includegraphics[width=0.9\linewidth]{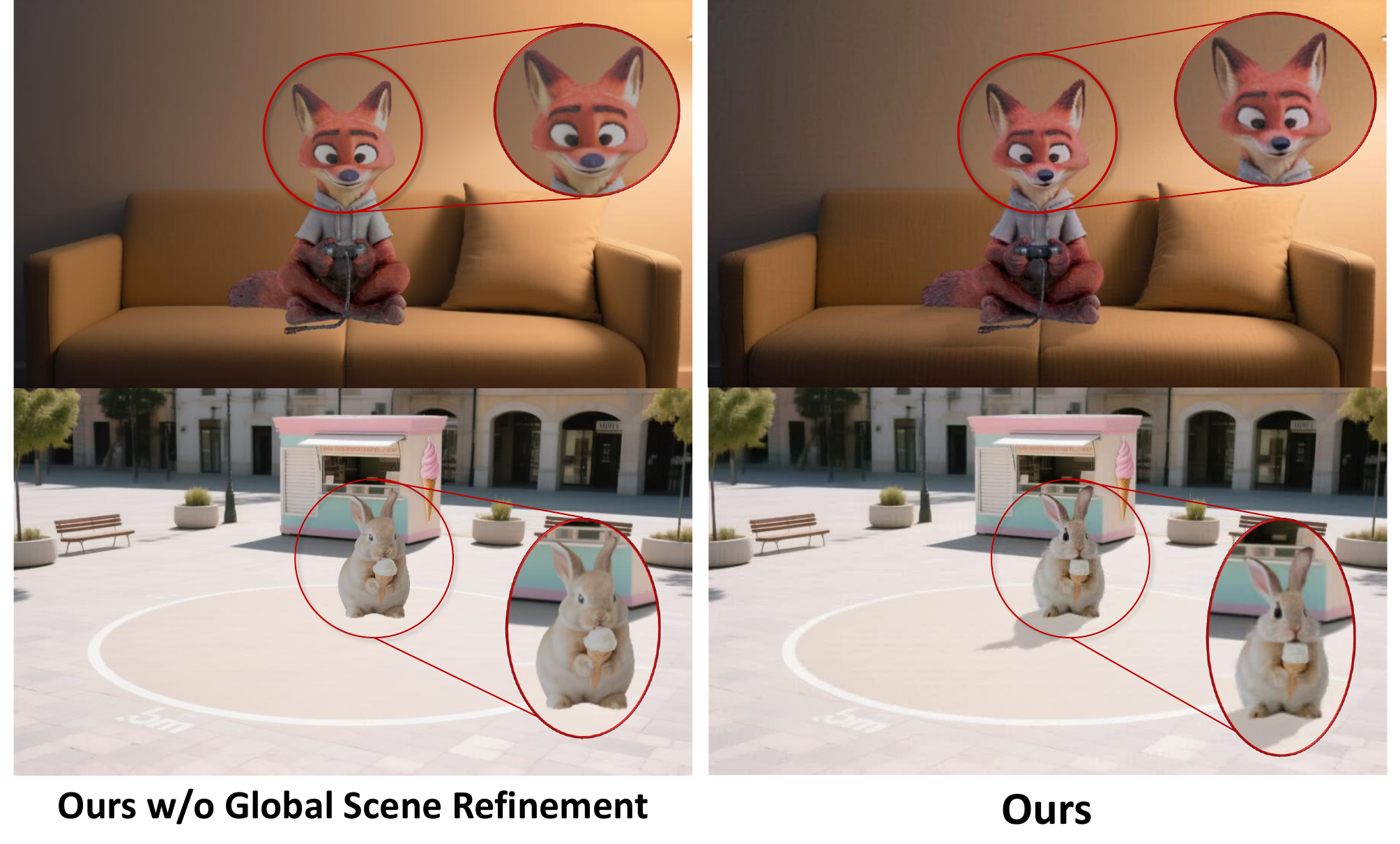}
    \vspace{-1em}
    \caption{Ablation study of our global scene refinement. We show that the refinement stage improves fine-grained appearance details and enhances the overall visual coherence.}
    \label{fig:analysis_refine}
    \vspace{-1em}
\end{figure}

\subsection{Ablation Study}
We conduct ablation studies to analyze how each component of Edit4D affects scene--character composition and 4D generation quality.
Tab.~\ref{tab:ablation} shows that removing initialization, contact loss, penetration loss, and global refinement degrades different metrics, while the full model achieves the best quantitative results.
Fig.~\ref{fig:analysis_composition} further visualizes the effect of each composition component.
Without initialization, the optimization often converges to an incorrect placement, whereas removing the contact loss causes the character to float above the support surface, and removing the penetration loss results in penetration into the scene geometry.
These results indicate that each constraint and refinement provide complementary benefits for robust placement, visual quality, and scene coherence.
Moreover, Fig.~\ref{fig:analysis_refine} demonstrates the effect of the global refinement stage. 
The refinement stage enhances fine-grained details while improving the overall visual coherence of the composed 4D scene.

\section{Conclusion}
In this work, we presented Edit4D, a framework for editable 4D scene generation that separates background scenes and dynamic characters for object-level control and composition.
Quantitative evaluations on VBench demonstrate state-of-the-art generation quality, while our editability evaluation shows that Edit4D better preserves unedited scene content and reduces editing time compared with holistic regeneration.
Qualitative results further demonstrate that Edit4D produces coherent 4D scenes and remains robust under large camera viewpoint changes, supporting reliable novel-view synthesis and dynamic camera control.
Our analysis validates the effectiveness of text-guided composition and contact-aware optimization for semantically and physically plausible character placement.
Together, these results show that Edit4D provides a practical and flexible framework for high-quality editable 4D scene generation.

{
\noindent\textbf{Acknowledgments}
This work was supported by Institute of Information communications Technology Planning Evaluation (IITP) grant funded by the Korea government (MSIT) (No.RS2023-00225630, Development of Artificial Intelligence for Text-based 3D Movie Generation).
\par}
{
    \small
    \bibliographystyle{ieeenat_fullname}
    \bibliography{main}
}

\clearpage
\maketitlesupplementary
\appendix
\section{Implementation details}
\label{sec:supp_impl}
We describe the implementation of Edit4D by summarizing each component: instruction decomposition, background scene generation, dynamic 3D character generation, scene--character composition, and global scene refinement.
Importantly, Edit4D is not tied to the specific pretrained models used in our implementation.
Each stage is modular and can be replaced by an alternative model with compatible input and output representations, allowing the framework to incorporate different open-source models without changing the overall pipeline.

\subsection{Instruction Decomposition}
We use GPT v5.2~\cite{openai_chatgpt} to decompose a user instruction into stage-specific prompts for background scene generation, character generation, motion generation, and scene--character composition. Each prompt contains only the information required by its corresponding stage, allowing the pipeline to operate automatically from a single user instruction while preventing unnecessary information from being propagated across stages.
The instruction template is summarized in Table~\ref{tab:instruction_decomposition}.

\subsection{Background Scene Generation}
Given a textual scene description, we first synthesize a frontal background scene image ($1024\times 1024$) using a text-to-image model~\cite{wu2025qwenimagetechnicalreport}.
To enable 3D reconstruction, we then generate a set of geometrically consistent multi-view images by extending the frontal view to additional camera poses via a multi-view generation framework.
The resulting multi-view set is used to estimate a 3D Gaussian splatting representation of the background.
In our implementation, we use VistaDream~\cite{wang2025vistadream} to jointly produce trajectory-conditioned multi-view inpainting images and reconstruct the background Gaussian splatting in a unified pipeline.

\subsection{Dynamic 3D Character Generation}
We generate a dynamic character in five steps: (1) image generation, (2) image-conditioned video generation, (3) 4D mesh estimation, (4) multi-view rendering, and (5) training dynamic Gaussian splatting.
We first synthesize a character image ($800\times800$) from a text prompt, and then generate an image-aligned motion video ($800\times800$, 60 frames) using an image- and text-conditioned video model~\cite{wan2025}.
The temporally consistent 4D mesh sequence is reconstructed using Motion 3-to-4~\cite{chen2026motion}, which estimates a canonical reference mesh and predicts per-frame vertex trajectories to recover temporally coherent geometry.

To render the character together with the Gaussian splatting background efficiently, we convert the 4D mesh to a dynamic Gaussian representation.
Specifically, we render multi-view images in Blender~\cite{Blender2024} using 30 cameras uniformly distributed on a sphere and bind Gaussian primitives to mesh faces following Mani-GS~\cite{gao2025mani}, so that Gaussians follow mesh deformations over time.
To further improve temporal stability under non-rigid deformation, we incorporate a lightweight deformation field following~\cite{yang2024deformable} that predicts per-time offsets of Gaussian position, rotation, scale, and color.
We execute all of the image and video generation steps in ComfyUI~\cite{comfyui}. 
Moreover, to improve generation quality, we additionally employ task-specific positive and negative prompts to suppress common generation artifacts (e.g., shadows, background leakage, and watermark-like artifacts).

\subsection{Scene--Character Composition}
Given the background and dynamic character Gaussians, we optimize a single global rigid transform of the character, parameterized by a global rotation and translation.
The rotation is applied around the character center, followed by translation, while the character scale is fixed to a predefined value between 0.6 and 1.2.
We initialize the character position using the bounding boxes of the character mesh and the target Gaussians.
Specifically, we first align the character center with the center of the target region.
Following the $y$-down coordinate convention, we then apply a small offset along the negative $y$-axis to place the character slightly above the support surface before optimization.
This initialization provides a stable starting point for subsequent contact-aware optimization.
Next, we minimize a weighted sum of contact and penetration losses for 1000 iterations:
\begin{align}
\mathcal{L}_{\text{composition}} =
\lambda_{\text{contact}}\mathcal{L}_{\text{contact}}
+
\lambda_{\text{penetration}}\mathcal{L}_{\text{penetration}}.
\end{align}
In our experiments, we set $\lambda_{\text{contact}}=10^2$ and $\lambda_{\text{penetration}}=10^4$, assigning a stronger penalty to interpenetration.
For the penetration loss, we precompute a signed distance field (SDF) from the character mesh on a $128^3$ grid with a padding of $0.05$, where negative SDF values indicate points inside the character mesh.
During optimization, scene points are transformed into the character coordinate system and queried against the SDF using trilinear interpolation.
Points with negative SDF values are penalized to discourage interpenetration.
The optimized global transform is then applied consistently to the dynamic character Gaussians across all frames.

\subsection{Global Scene Refinement}
During the refinement, we jointly optimize the background scene, the canonical character Gaussian representation, and the deformation network for 4,000 iterations.
During the first 500 iterations, we optimize the canonical frame using multi-view renderings sampled from the camera trajectories generated during the background scene generation stage.
Also, during the remaining 3,500 iterations, we randomly sample either the canonical frame or a dynamic frame for optimization. 
The dynamic frames are optimized only from the frontal view to preserve character identity and temporal consistency.

For shadows, we first estimate frame-wise HDR environment maps from 25 background views using LuxDiT~\cite{liang2026luxdit}.
The per-view lighting estimates are aligned to a common world coordinate system using the corresponding camera rotations and aggregated using an element-wise median:
\begin{equation}
L(\omega) =
\operatorname*{median}_{k=1,\ldots,25} L_k(\omega),
\end{equation}
where $\omega\in\mathbb{S}^{2}$ denotes a light direction and $L_k(\omega)=[R_k(\omega),G_k(\omega),B_k(\omega)]$ is the RGB radiance predicted for that direction in view $k$. 
This aggregation reduces the view-dependent inconsistencies while preserving lighting that is consistent across views.
Given the linear HDR radiance, we compute its relative luminance $Y(\omega)$ as follows:
\begin{equation}
Y(\omega)
=
0.2126R(\omega)+0.7152G(\omega)+0.0722B(\omega),
\end{equation}
We select high-luminance candidate regions from directions within the top $10\%$ of the relative luminance distribution and further retain salient localized lighting regions.
The selected regions are spatially smoothed to construct the directional environment map $L_{\mathrm{dir}}$.
Given $L_{\mathrm{dir}}$, we sample $N=512$ light directions by importance sampling according to their luminance.
For each environment map pixel $i$, the sampling probability is defined following:
\begin{equation}
p_i =
\frac{Y_{\mathrm{dir},i}\Delta\Omega_i}
{\sum_j Y_{\mathrm{dir},j}\Delta\Omega_j},
\end{equation}
where $Y_{\mathrm{dir},i}$ and $\Delta\Omega_i$ denote the luminance and solid angle associated with pixel $i$, respectively. 
The full environment map $L$ is sampled in the same manner to estimate $E_{\mathrm{all}}$.

To obtain shadow receivers, we unproject visible background pixels into 3D world space using the rendered depth and obtain their surface normals from multiple depth observations.
For each receiver point $\mathbf{x}_\text{r}$ with normal $\mathbf{n}_\text{r}$, we trace rays toward the sampled light directions following Gaussian ray tracing~\cite{3dgrt2024, gu2024gtracer}. 
For each sampled direction $\omega_i$, it returns the accumulated Gaussian opacity $\alpha(\mathbf{x}_\text{r},\omega_i)\in[0,1]$, representing the fraction of light blocked by the character along the ray.
Let $Y_\text{dir}$ denote the luminance of $L_{\text{dir}}$. 
Following the standard illumination integral used
in physically based inverse rendering~\cite{wu2023nefii}, we define the irradiance as follows:
\begin{align}
E_{\text{all}}(\mathbf{x}_\text{r})
&=
\int_{\mathbb{S}^2}
Y(\omega)
\max(\mathbf{n}_\text{r}^\top\omega,0)\,d\omega,
\nonumber \\
E_{\text{blocked}}(\mathbf{x}_\text{r})
&=
\int_{\mathbb{S}^2}
\alpha(\mathbf{x}_\text{r},\omega)
Y_{\text{dir}}(\omega)
\max(\mathbf{n}_\text{r}^\top\omega,0)\,d\omega,
\end{align}
In practice, we approximate both integrals using Monte Carlo estimation with importance samples drawn from $L$ and $L_{\mathrm{dir}}$, respectively.
Then, the relative illumination loss is as below:
\begin{equation}
r_{\text{loss}} = \frac{L_{\text{blocked}}}{E_{\text{all}}+\epsilon},
\end{equation}
We apply this loss only to the background contribution in linear RGB.
For an image pixel $p$ corresponding to receiver $\mathbf{x}_r$, the final composite is as follow:
\begin{equation}
\mathbf{C}^{\text{out}}_t(p) = \mathbf{C}^{\text{comp}}_t(p) -
w^{\text{bg}}_t(p)\,
r_{\text{loss},t}(\mathbf{x}_r)\,
\mathbf{C}^{\text{bg}}(p).
\end{equation}
where $\mathbf{C}^{\text{comp}}_t$ is the unshadowed linear RGB composite and $w^{\text{bg}}_t$ is the soft background contribution, which prevents the attenuation from being applied to the character.
The fused environment maps $L$ and $L_{\text{dir}}$ are fixed across the sequence, while the character dependent opacity
$\alpha_t(\mathbf{x}_r,\omega)$ is recomputed at every frame.
Consequently, the resulting shadows follow the character motion under temporally consistent background illumination.

Throughout the refinement, we optimize the objective in Eq.~(10), which combines latent-space and image-space consistency losses with $\lambda_1=1$ and $\lambda_2=0.25$. 
We use Stable Diffusion v2.1~\cite{LDM} as the diffusion prior.
All final 4D scenes are rendered at a resolution of $900 \times 512$.

\subsection{Runtime}
On a single RTX Pro 6000 Blackwell GPU, Edit4D takes on average approximately 2 hours per example: 10 minutes for static scene generation, 70 minutes for dynamic character generation, 1 minute for composition, and 45 minutes for global scene refinement.
This staged design enables component-level editing of the background scene, dynamic character, and their composition, which is difficult in holistic video-to-4D pipelines.

We also compare the runtime of prior text-to-4D methods with publicly available code when generating the same number of frames as Edit4D.
4D-fy and Dream-in-4D typically require more than 24 hours per example due to expensive computation of radiance field and score distillation sampling-based optimization, while Free4D takes approximately 3 hours when including multi-view video generation and 4D Gaussian optimization.
More importantly, when editing an existing scene, holistic 4D generation pipelines generally require regenerating the entire 4D result from scratch.
Edit4D can instead reuse pre-generated background and character assets, requiring only scene-character composition and optional global scene refinement.
This substantially reduces the cost of iterative editing and makes Edit4D a more practical framework for controllable 4D scene creation.

\begin{table*}[t]
\centering
\small
\begin{tabular}{p{0.21\linewidth} p{0.73\linewidth}}
\toprule
\textbf{Task} & \textbf{Prompt guideline used for Instruction Decomposition} \\
\midrule

\textbf{Background Scene Text}
&
\textbf{Objective:}
Extract only the background scene information from the user instruction.

\textbf{Rules:}
(i) Include only scene, environment, and objects.
(ii) Exclude character identity, appearance, motion, and placement.
(iii) Do not infer or introduce any information that is not explicitly mentioned.
\\[0.8em]

\midrule

\textbf{Character Text}
&
\textbf{Objective:}
Extract only the character identity and appearance.

\textbf{Rules:}
(i) Exclude scene, background, motion, pose, and placement information.
(ii) Do not infer clothing, accessories, facial features, or other attributes unless explicitly specified.
\\[0.8em]

\midrule

\textbf{Motion Text}
&
\textbf{Objective:}
Extract only the character motion or action.

\textbf{Rules:}
(i) Exclude appearance, scene, and placement information.
(ii) Describe only the character motion.
(iii) Do not introduce additional actions or pose details.
\\[0.8em]

\midrule

\textbf{Composition Text}
&
\textbf{Objective:}
Extract only the target object or region for scene--character composition.

\textbf{Rules:}
(i) Return a concise noun or noun phrase.
(ii) Do not infer spatial layouts or additional relationships.
\\

\bottomrule
\end{tabular}
\caption{Prompt guidelines used by GPT~\cite{openai_chatgpt} for instruction decomposition. A user instruction is decomposed into stage-specific texts for the background scene, character, motion, and composition.}
\label{tab:instruction_decomposition}
\vspace{-0.5em}
\end{table*}

\section{Baseline and Datasets}
\begin{table*}[t!]
    \centering
    \footnotesize
    \setlength{\tabcolsep}{2.5pt}
    \renewcommand{\arraystretch}{0.95}
    \begin{tabular}{
        >{\centering\arraybackslash}m{0.25\linewidth}|
        >{\centering\arraybackslash}m{0.31\linewidth}|
        >{\centering\arraybackslash}m{0.25\linewidth}|
        >{\centering\arraybackslash}m{0.14\linewidth}
    }
        \toprule
        \makecell[c]{Background Scene Text} &
        \makecell[c]{Character Text} &
        \makecell[c]{Motion Text} &
        \makecell[c]{Composition Text} \\
        \bottomrule
        Soft-serve counter with a pastel menu board, stacked waffle cones, napkin dispenser, sealed syrup bottles, clean white walls, and studio lighting & Baby panda holding a small vanilla ice-cream cone with both paws &  Gently licking a vanilla ice-cream cone while holding it with both paws, with tiny head bob & Counter \\
        \midrule
        Cel-shaded reading corner with a floor cushion, small table, closed books, bookshelf, pastel wall, and flat even lighting & Anime-style baby panda sitting while holding and reading an open book with both paws & Sitting and reading an open book, with blink and small head tilt & Book on the table \\
        \midrule
        Minimal lounge with a beige fabric sofa, soft cushion, neutral wall, light rug, wooden sofa legs, and warm natural indoor lighting & Ring-tailed lemur holding a clear plastic boba cup & Holding a boba cup and sipping through a wide straw, with mouth and straw motion & Sofa \\
        \midrule
        Bright daytime plaza with light stone pavers, a pastel gelato kiosk, benches, planters & Baby rabbit holding a small vanilla ice-cream cone with both paws & Gently licking a vanilla ice-cream cone while holding it with both paws, with tiny head bob & Yard \\
        \midrule
        Sound-treated rehearsal studio with matte wood floor, acoustic wall panels, bass traps, cable hooks, powered-off amps, soft practical lighting, and a clear floor area & Cat seated upright, singing with an open mouth while holding a handheld microphone & Singing into a handheld microphone with subtle head movement & Center stage \\
        \midrule
        Modern indoor atrium at dusk with lounge chairs, a small side table, planter boxes, stone floor, wood accents, soft ceiling lights, and an empty circular stone landing pad & Square stone fire pit with stacked gray stones, burning firewood logs, tall orange flames, and glowing embers & The flames flicker naturally upward and the ember glow changes subtly & Landing pad \\
        \midrule
        Open meadow with short even grass, distant hill horizon, neutral daylight, and a round flat pad at the center & White goat with curved horns and a small beard, standing while lowering its head to sip beer from a clear glass & Lowering its head to drink pale beer from a glass & Flat pad \\
        \midrule
        Modern kitchen with a front-facing island, light wood countertop, white cabinets, matte tile backsplash, utensils, towel, and fruit bowl & Kangaroo wearing a white chef hat, holding a black frying pan with a sunny-side-up egg and a wooden spatula & Holding a frying pan and wooden spatula, gently tilting the pan while moving the spatula across the pan in a cooking motion & Floor \\
        \midrule
        Sunny park lawn with a red-white gingham picnic blanket, an empty dessert plate at the center, a closed wicker basket, folded napkins & Baby monkey holding a partially unwrapped plain candy bar with both hands & Taking small bites from a partially unwrapped candy bar, with natural chewing and head bob & Left-side of blanket \\
        \midrule
        Warm wooden treehouse study nook with a round porthole window, low desk, closed books, floor cushion, powered-off string lights, shelf, and pastel wood tones & Stylized 3D monster with small horns, big expressive eyes, a soft plush-like surface, sitting cross-legged while holding & Sitting cross-legged and reading an open book, with gentle head tilt and blink & Cushion \\
        \midrule
        Backyard garden with a raised wooden planter, groomed dark soil, rows of carrot greens and herbs, soft overcast daylight, and a circular clear patch of soil at the center of the bed & Rabbit holding a bright orange carrot & Holding and eating a bright orange carrot, with head bob & Center of ground \\
        \midrule
        Urban block corner with asphalt road, crosswalk paint, concrete sidewalk, a clearly visible vacant gravel lot, low curb lines, an empty center area for placement & Mid-rise glass, concrete, and steel building with exterior flames wrapping along the outside walls and window frames & Exterior flames licking up the outside walls and window frames of a building, with hot spots and mild char on the outer surface & Vacant lot \\
        \midrule
       A beige fabric sofa with a soft cushion, soft and warm indoor lighting, cozy living room & Disney-style anthropomorphic red fox character wearing a light gray hoodie, sitting while holding a game controller & Playing a video game while sitting cross-legged and holding a game controller & Sofa \\
        \bottomrule
    \end{tabular}
    \caption{Text prompt used in our evaluation for each stage. We report the background scene, character, motion, and composition prompts in Edit4D.}
    \label{tab:text}
\end{table*}

Our goal is text-to-4D scene generation with editability.
We compare our method with four baselines, including two object-centric generation methods~\cite{bahmani20244dfy, zheng2024unified} and two general text-to-4D methods~\cite{yu20244real, liu2025free4d}.
For pairwise comparison, we follow the evaluation protocol of Free4D~\cite{liu2025free4d} and adopt the three experimental settings provided on its project page.
Since the original prompts are not publicly available for most baseline results, we manually construct user instructions that closely match their publicly released visual content.
Following Free4D, we evaluate 7, 6, and 6 samples for settings (a), (b), and (c), respectively.
Because several samples are shared across settings, this corresponds to 13 unique instructions in total.
Tab.~\ref{tab:text} provides all stage specific prompts for background, character, motion, and composition used in our evaluation.
\section{Capability}
\begin{table*}[!t]
    \renewcommand{\arraystretch}{1.5}
    \centering
    \resizebox{\linewidth}{!}{\begin{tabular}{lccccc}
        \toprule
        Feature & 4Real~\cite{yu20244real} & Dream-in-4D~\cite{zheng2024unified} & 4Dfy~\cite{bahmani20244dfy} & Free4D~\cite{liu2025free4d} & Edit4D (Ours) \\
        \midrule
        Input & Text, Video & Text & Text & Text, Video & Text\\
         Output Representation & \makecell{Dynamic Gaussian \\ splatting  for whole scene\\} & \makecell{Dynamic NeRF \\ for whole scene\\} & \makecell{Dynamic Gaussian \\ splatting for whole scene\\} & \makecell{Dynamic Gaussian \\ splatting for whole scene\\} & \makecell{Background Scene \& Dynamic \\ Character Gaussian splatting\\} \\
        Scene-Character Separation & \ding{55}  &  \ding{55} &
\ding{55} & \ding{55} & \ding{51}   \\ 
        Component Editing & \ding{55}  &  \ding{55} &
\ding{55} & \ding{55} & \ding{51}   \\
        Character Composition &  \ding{55}  &  \ding{55} &
\ding{55} & \ding{55} & \ding{51}   \\
        \bottomrule
    \end{tabular}}
    \vspace{-0.5em}
    \caption{Capability comparison of previous 4D generation pipelines with Edit4D. Edit4D supports explicit scene-character separation, component-level editing, and character composition.
    }
    \vspace{-0.5em}
    \label{tab:comparison}
\end{table*}

In Table~\ref{tab:comparison}, we compare Edit4D with existing text-to-4D generation methods in terms of their capabilities.
Specifically, we compare representative scene-level and object-level methods with respect to input modality, output representation, scene--character separation, component-level editing, and character composition.
Existing methods typically generate a holistic dynamic scene representation in which the background and dynamic characters are tightly coupled.
In contrast, Edit4D represents the background scene and dynamic characters as separate Gaussian representations, enabling independent component-level editing and flexible scene--character composition.
\begin{figure}[!t]
    \centering
    \includegraphics[width=0.9\linewidth]{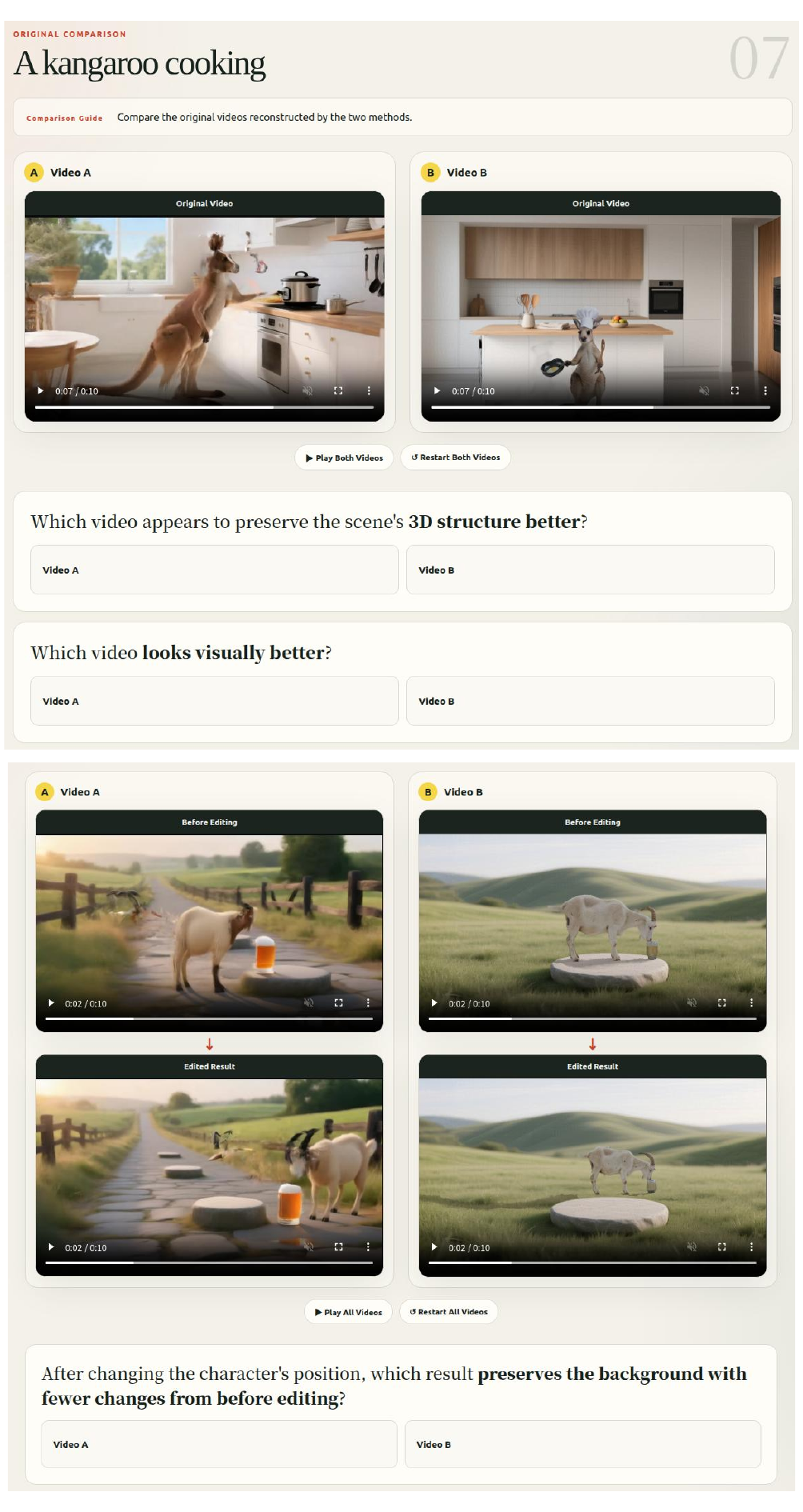}
    \caption{User study interface. Participants evaluate Edit4D and Free4D based on 3D structure and visual quality for the original 4D scenes, as well as background preservation after editing.}
    \vspace{-1em}
    \label{fig:user_study}
\end{figure}

\section{Details on User Study}
\label{sec:supp_user_study}
We conducted an anonymous pairwise user study comparing Edit4D with Free4D~\cite{liu2025free4d}.
A total of 25 participants evaluated the rendered 4D scenes through a web-based interface, where the two methods were presented as Video A and Video B without revealing the method names.
The scene order and the left--right placement of the two methods were randomized to reduce presentation bias.
For a fair comparison, we regenerated the Free4D results using its publicly available implementation.
Since Free4D requires a reference video, we generated the corresponding inputs using the same Qwen~\cite{qwen3technicalreport} and WAN~\cite{wan2025} based generation pipeline used for Edit4D, with the same video resolution and frame setting.
Both methods were rendered from multiple viewpoints for comparison.
For the original 4D scenes, participants compared 13 scenes and answered two questions regarding 3D Structure and Visual Quality.
Specifically, participants selected which result better preserved the perceived 3D structure of the scene and which result appeared visually better.
We additionally evaluated editability using five scenes for both scene--character recomposition and character appearance editing.
For each method, participants were shown the result before and after editing and asked which method better preserved the background content after the edit.
An example of the evaluation interface is shown in Fig.~\ref{fig:user_study}.

\begin{figure*}[!t]
    \centering
    \vspace{2em}
    \includegraphics[width=0.98\linewidth]{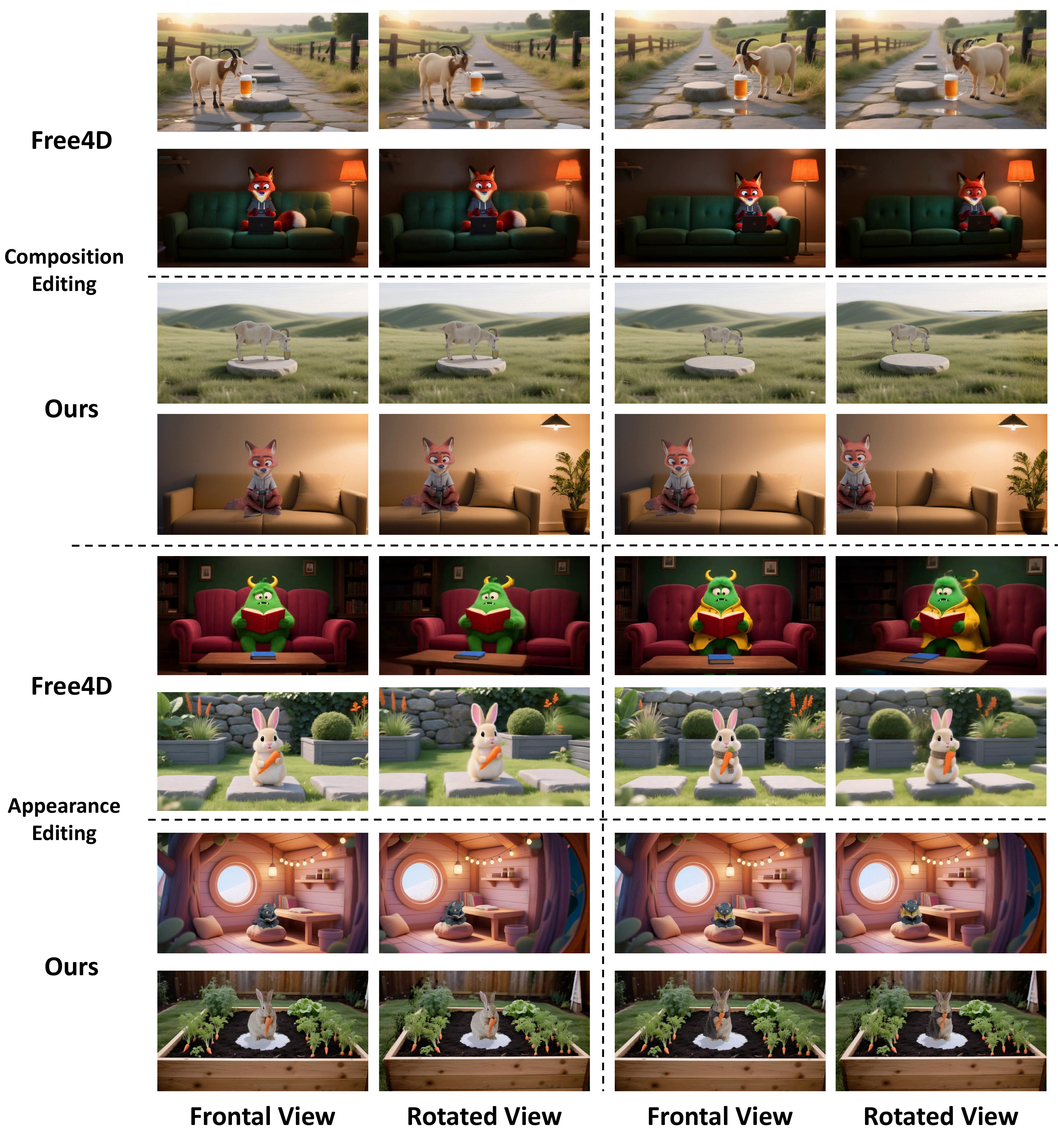}
    \caption{Qualitative examples used in the editability evaluation. We compare Free4D and Edit4D for scene--character recomposition and character appearance editing, showing both frontal and rotated views. 
    Free4D requires full scene regeneration for each edit, whereas Edit4D reuses the existing scene assets and better preserves the unedited background content across viewpoints.}
    \label{fig:editing_evaluation_supp}
    \vspace{2em}
\end{figure*}

\section{Evaluation on Editability}
We further evaluate the editability of Edit4D through character appearance editing and scene--character recomposition.
We compare against Free4D, which, to our knowledge, is the only scene-level text-to-4D generation method with publicly available source code, allowing us to directly reproduce its pipeline for controlled editing comparisons.
Since Free4D represents the scene holistically, each edit requires full regeneration of the 4D scene.
For a fair comparison, we generate the edited reference videos using the same video generation model as Edit4D and run Free4D using its publicly released implementation and settings.

Our evaluation focuses on how well the unedited background is preserved after editing.
We evaluate five character scenes for each of the two editing tasks and compare the original and edited renderings across multiple camera viewpoints and time steps shown in Fig.~\ref{fig:editing_evaluation_supp}.
To prevent changes in character position, scale, or appearance from affecting the evaluation, we localize the character using Grounding DINO~\cite{liu2023grounding} and apply a shared mask covering the union of the character regions in the original and edited images.
Thus, the evaluation measures the consistency of the background outside the edited region.

We assess editing success using three VLMs, GLM-4.6~\cite{vteam2025glm45vglm41vthinkingversatilemultimodal}, Qwen3-VL~\cite{qwen3technicalreport}, and Mistral-3.1~\cite{mistralsmall3124binstruct2503}, based on whether the original background scene is preserved after editing.
Their decisions are aggregated across sampled viewpoints and time steps.
We separately measure background preservation using CLIP~\cite{radford2021learning} and DINOv2~\cite{oquab2023dinov2} similarities computed between the masked original and edited images, where higher similarity indicates better preservation of the unedited background.
As reported in the main paper, Edit4D consistently achieves higher editing success and background preservation than Free4D for both editing tasks while requiring substantially less editing time.
\begin{figure}[!t]
    \centering
    \includegraphics[width=\linewidth]{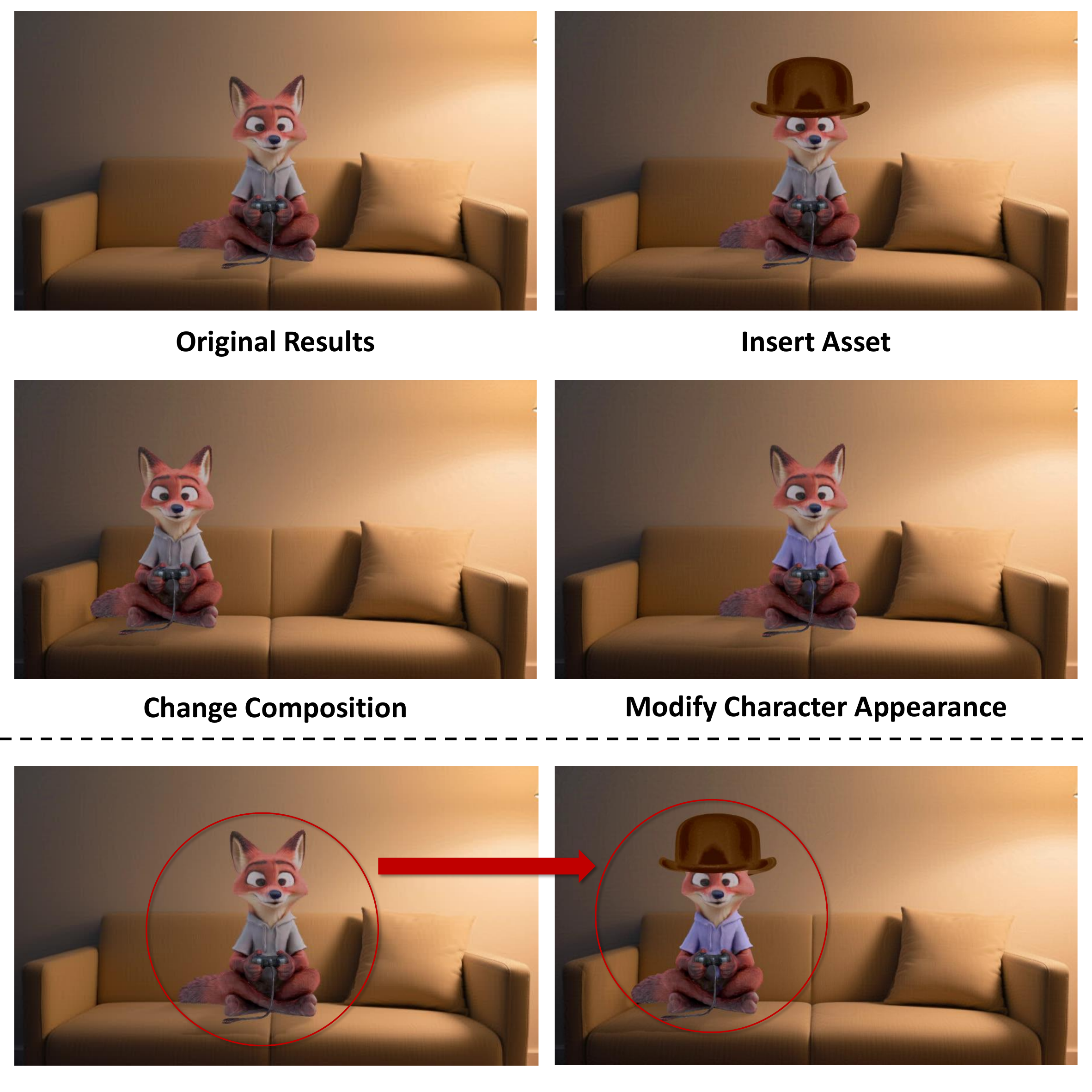}
    \caption{Object-level editing enabled by Edit4D. Our modular 4D representation supports character appearance modification, asset insertion, and scene--character recomposition without regenerating the entire scene.}
    \label{fig:editing}
\end{figure}

\begin{figure}[!t]
\centering
\setlength{\tabcolsep}{1.5pt}
\renewcommand{\arraystretch}{1.0}
\begin{tabular}{cc}
\includegraphics[width=0.485\linewidth]{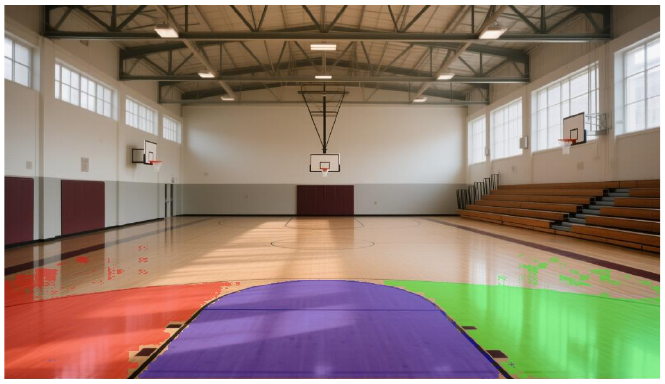} &
\includegraphics[width=0.485\linewidth]{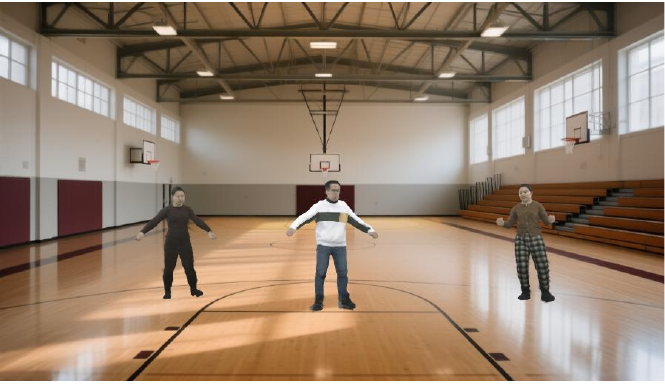} \\[-1.2mm]
\multicolumn{2}{c}{
\phantomsubcaption\label{fig:additional_results_a}
\footnotesize (a) Composition of multiple characters
} \\[1.2mm]
\includegraphics[width=0.48\linewidth]{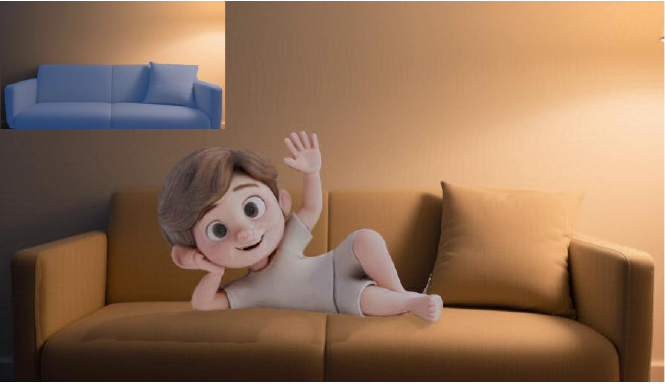} &
\includegraphics[width=0.48\linewidth]{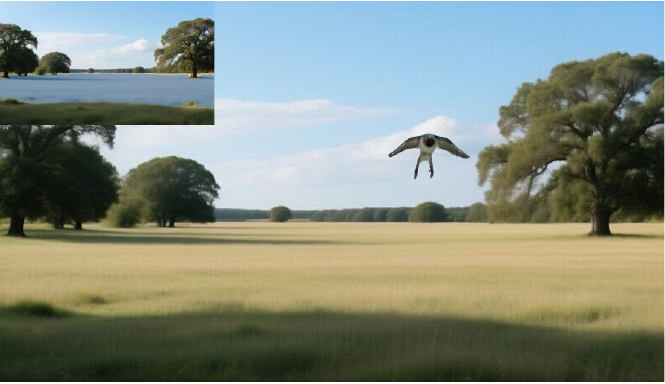} \\[-1.2mm]
\phantomsubcaption\label{fig:additional_results_b}
\footnotesize (b) Lying character &
\phantomsubcaption\label{fig:additional_results_c}
\footnotesize (c) Flying bird \\
\end{tabular}
\vspace{-0.5em}
\caption{
Additional results of our pipeline.
We segment target regions from text and align the character via the proposed composition algorithm.
}
\vspace{-0.5em}
\label{fig:additional_results}
\end{figure}

\section{Additional qualitative results}
We provide additional qualitative results to further demonstrate the editing capability and flexibility of Edit4D.
As shown in Fig.~\ref{fig:editing}, Edit4D supports asset insertion, scene--character recomposition, and character appearance modification using the scene presented in the main paper.
In the asset insertion example, we manually place a hat represented as a Gaussian Splatting asset onto the character.
Scene--character recomposition is performed by specifying ``left sofa'' as the target region and repositioning the character through our composition framework.
To modify the character appearance, we construct a 3D mask for ``cloth'' using our masking method and update the color values within the masked region.
Since the background and character are represented as reusable 3D assets, these edits can be performed while preserving the remaining scene content without regenerating the entire 4D scene.

We further demonstrate the flexibility of our scene--character composition framework in Fig.~\ref{fig:additional_results}.
As shown in Fig.~\ref{fig:additional_results_a}, our framework can incorporate alternative dynamic character representations.
Specifically, we integrate Animatable Gaussians~\cite{li2024animatablegaussians} and obtain plausible multi-character compositions, while visualizing left, center, and right floor masks for controllable placement.
To maximize compatibility with diverse character representations, our main pipeline adopts a video-to-mesh model that does not rely on explicit rigging or skeleton annotations.

Fig.~\ref{fig:additional_results_b} further presents a more challenging interaction with a large contact area between the character and the scene.
For non-contact interactions such as hovering or flying, we replace the contact objective with a distance-thresholded variant that maintains a target distance from the scene surface.
As shown in Fig.~\ref{fig:additional_results_c}, this produces stable and non-intersecting floating characters within the same optimization framework.
Together, these results demonstrate that the proposed composition framework can accommodate diverse character representations, contact configurations, and spatial relationships.

\begin{figure*}[!t]
\centering
\includegraphics[
    width=0.99\linewidth,
]{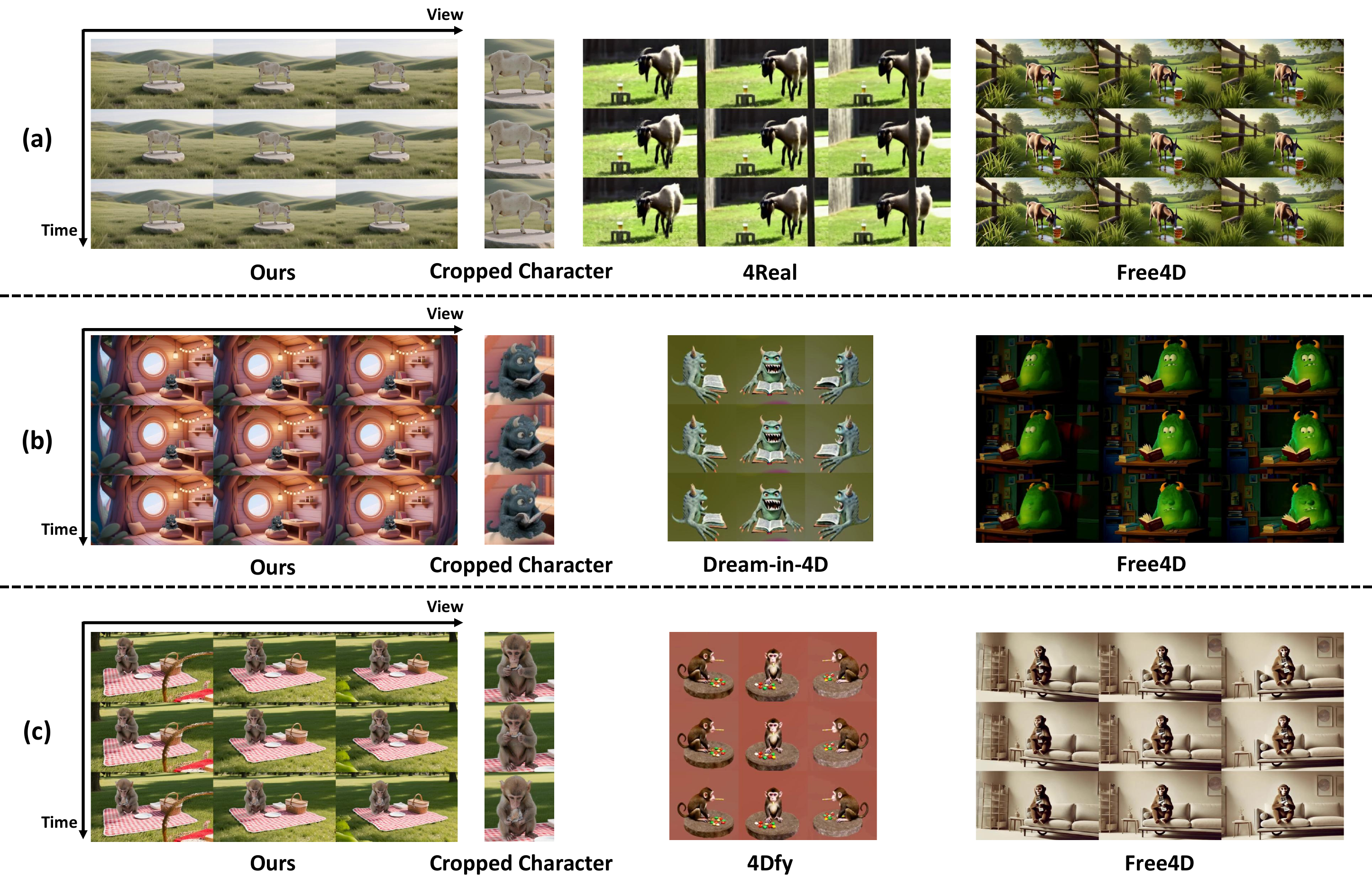}
\caption{Additional Qualitative comparison of our method with 4Real~\cite{yu20244real}, Dream-in-4D~\cite{zheng2024unified}, 4Dfy~\cite{bahmani20244dfy}, and Free4D~\cite{liu2025free4d}. We visualize samples from experimental settings (a), (b), and (c) in the main paper.}
\label{fig:qualitative_supple}
\end{figure*}

\begin{figure*}[!t]
\centering
\includegraphics[
    width=0.95\linewidth
]{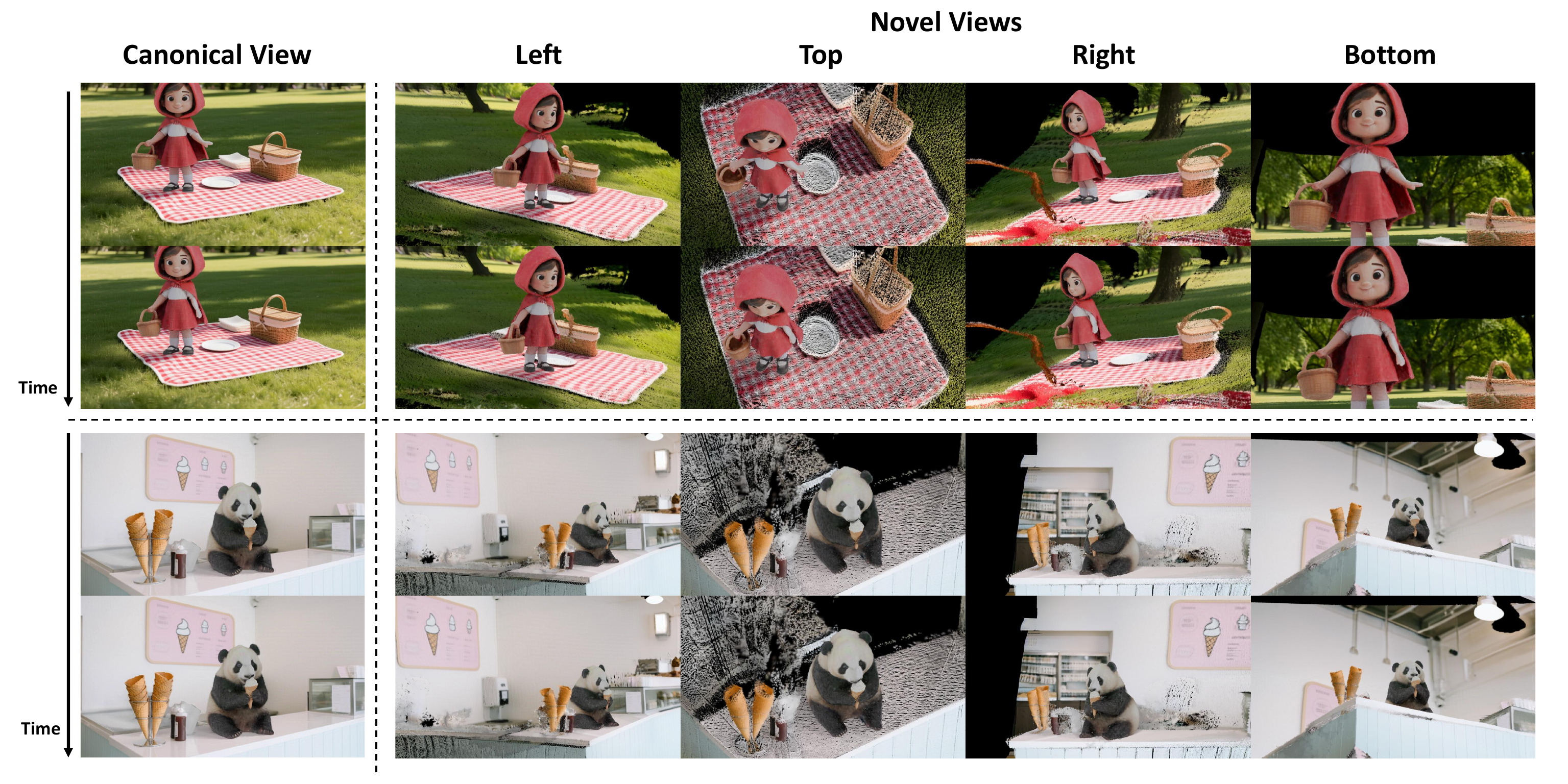}
\caption{Additional visualization of generated 4D scenes from the frontal view and four diverse novel camera views over time.}
\label{fig:novel_view_rendering_supple}
\end{figure*}

In addition, we present additional generation results across diverse text prompts in Fig.~\ref{fig:qualitative_supple}, showing that Edit4D consistently produces coherent and visually plausible 4D scenes.
Furthermore, Fig.~\ref{fig:novel_view_rendering_supple} provides additional novel-view renderings, demonstrating that our method maintains high visual quality under large camera viewpoint changes.

\section{Analysis on Contact Map}
\begin{figure}[!t]
    \centering
    \includegraphics[width=\linewidth]{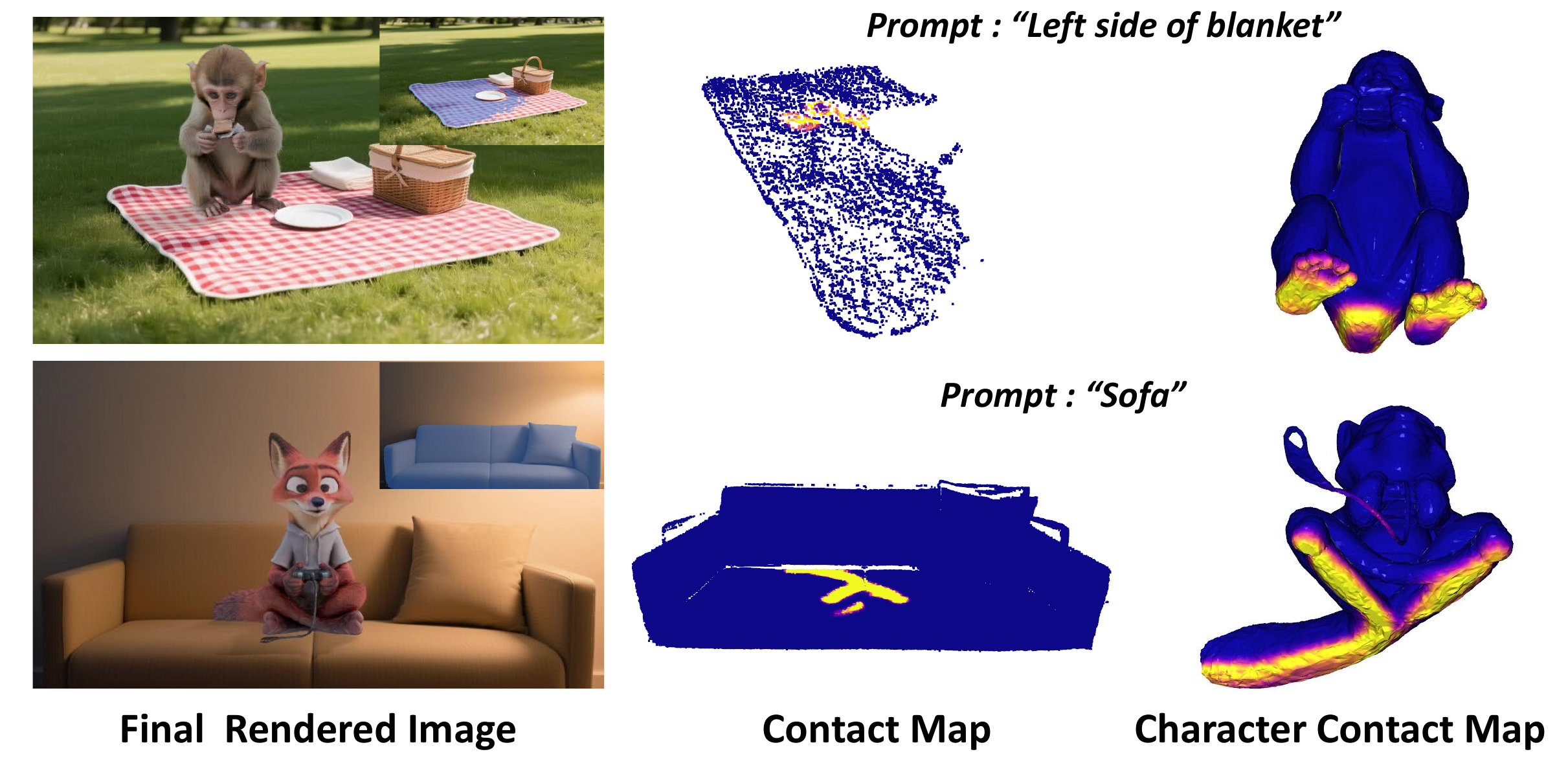}
    \caption{Rendered contact map from 3D visualizer. 
    Black and blue indicate the scene object and character, respectively, while brighter yellow represents higher contact intensity. We also visualize the target object mask in the figure.}
    \label{fig:analysis_contact}
\end{figure}

We visualize the contact intensity map produced by our composition stage in Fig.~\ref{fig:analysis_contact}.
Given a composition text, Edit4D first localizes the corresponding target region in the generated background scene and then optimizes the character placement using contact-aware constraints.
The high-contact regions align well with the target area specified by the composition text, showing that Edit4D can localize target placement regions and establish plausible character-scene contact.
In addition, the character contact maps show that the contact response is mainly activated around physically plausible support regions, such as the lower body or contact-facing surface of the character.
Overall, these analyses demonstrate that Edit4D enables physically plausible scene--character composition while maintaining the editability and rendering quality of the generated 4D scene.

\begin{figure}[!t]
\centering
\includegraphics[
    width=0.95\linewidth
]{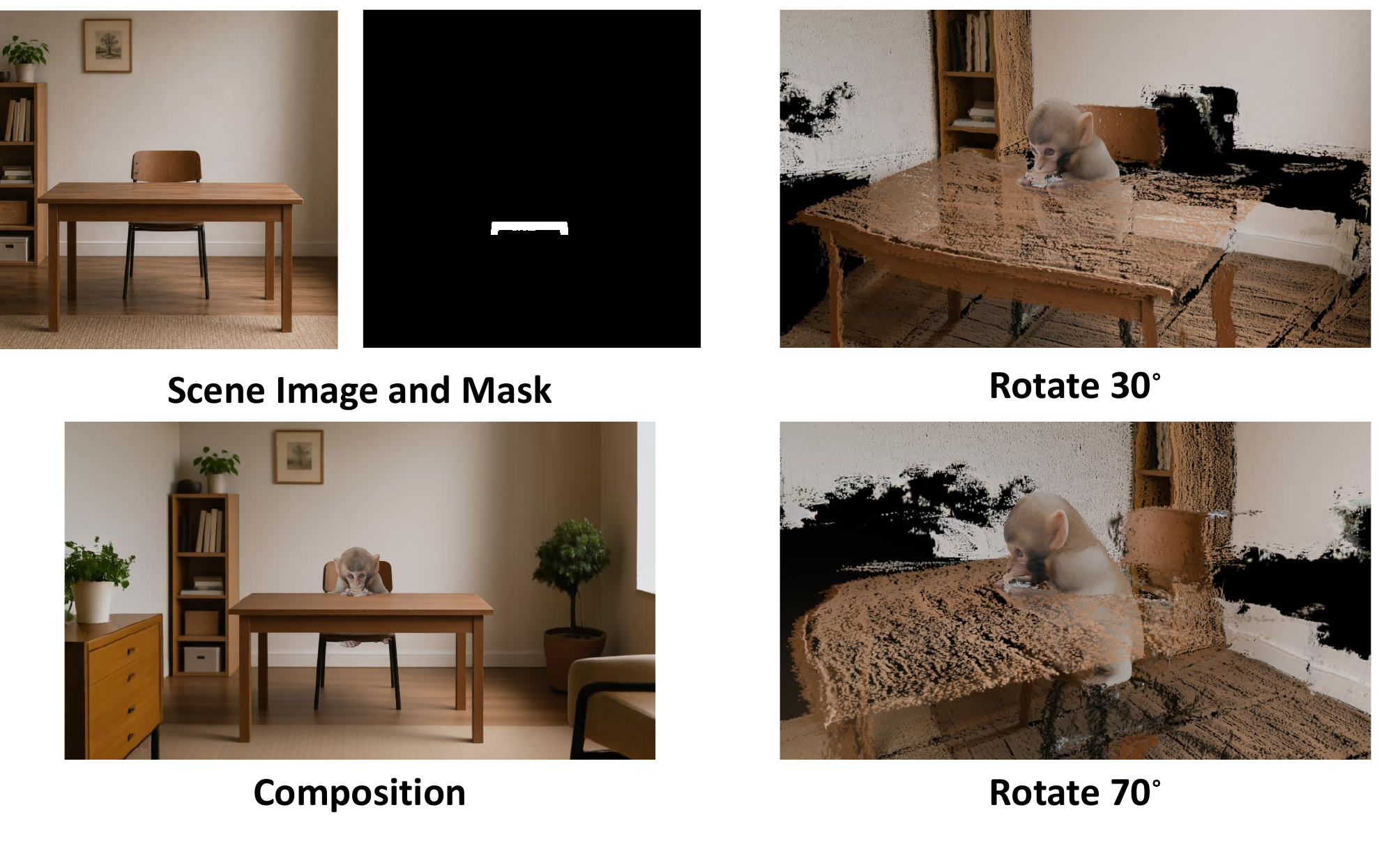}
\caption{Failure case caused by an occluded composition target.
The partially hidden chair produces an incomplete target region, leading to inaccurate character placement and intersection with the desk geometry in rotated views.}
\label{fig:limitation}
\end{figure}
\section{Limitations}
Our scene--character composition relies on text-guided segmentation to identify the target region for character placement.
Therefore, the target object should be sufficiently visible and spatially localized in the input scene.
Fig.~\ref{fig:limitation} shows a representative failure case in which the target chair is partially occluded by the desk.
Although the visible chair region can be identified, the incomplete target region leads to inaccurate character placement and intersection with the surrounding desk geometry.
This limitation could be alleviated by incorporating stronger 3D scene understanding or occlusion-aware target completion into the composition process.

Our pipeline also relies on several pretrained modules for intermediate predictions.
Errors from these components can propagate to the final 4D scene.
For example, imperfect dynamic character reconstruction may introduce temporal or geometric inconsistencies, while inaccurate text-guided segmentation can lead to errors in target localization and composition.
Similarly, artifacts in the generated image or reference video can affect the quality of the reconstructed scene or character.
Although our refinement stage can alleviate moderate inconsistencies, severe upstream errors may still remain in the final result.
Improving robustness to such intermediate prediction errors remains an important direction for future work.
\section{Future Work}
Edit4D opens several directions for extending editable 4D scene generation.
One promising direction is to incorporate more diverse editable assets into the component-wise representation.
Since Edit4D constructs background scenes and dynamic characters as separate 3D components, these assets can be inserted or reused across different compositions.
Another useful extension is interactive control over scene--character composition, where users can refine placement, contact regions, and spatial relationships after generation.
Improving target localization under occlusion through stronger 3D scene understanding or target completion could further increase the robustness of composition.
In addition, reducing sensitivity to errors in text-guided segmentation, dynamic character reconstruction, and generated reference images or videos would improve consistency in challenging cases.
These directions would further connect text-driven 4D generation with practical graphics-based editing workflows.

\end{document}